\documentclass[preprint,12pt,authoryear]{elsarticle}

\usepackage{amssymb}
\usepackage{amsmath}
\usepackage{amssymb}  
\usepackage[table]{xcolor}
\usepackage{array,multirow,float}
\usepackage{pifont}
\usepackage[colorlinks,linkcolor=blue,citecolor=blue,urlcolor=blue]{hyperref}

\newcommand{\cmark}{\ding{51}}%
\newcommand{\xmark}{\ding{55}}%

\usepackage{rotating} 
\newcommand{\rb}[1]{
  \rotatebox{90}{#1}
}

\def\secref#1{Sec.~\ref{#1}}
\def\figref#1{Fig.~\ref{#1}}
\def\tabref#1{Tab.~\ref{#1}}
\def\eqref#1{Eq.~(\ref{#1})}

\journal{Computer Vision and Image Understanding}

\begin{document}

\begin{frontmatter}



\title{Point-Plane Projections for Accurate LiDAR Semantic Segmentation in Small Data Scenarios} 


\author{Simone Mosco, Daniel Fusaro, Wanmeng Li, Emanuele Menegatti, Alberto Pretto} 
\ead{\{moscosimon, fusarodani, liwanmeng, emg, alberto.pretto\}@dei.unipd.it}

\affiliation{organization={Department of Information Engineering, University of Padova, Italy},
            addressline={Via Gradenigo 6/b}, 
            city={Padova},
            postcode={35131}, 
            state={Italy},
            country={IT}}

\begin{abstract}
LiDAR point cloud semantic segmentation is essential for interpreting 3D environments in applications such as autonomous driving and robotics. Recent methods achieve strong performance by exploiting different point cloud representations or incorporating data from other sensors, such as cameras or external datasets. However, these approaches often suffer from high computational complexity and require large amounts of training data, limiting their generalization in data-scarce scenarios. In this paper, we improve the performance of point-based methods by effectively learning features from 2D representations through point-plane projections, enabling the extraction of complementary information while relying solely on LiDAR data. Additionally, we introduce a geometry-aware technique for data augmentation that aligns with LiDAR sensor properties and mitigates class imbalance.
We implemented and evaluated our method that applies point-plane projections onto multiple informative 2D representations of the point cloud. Experiments demonstrate that this approach leads to significant improvements in limited-data scenarios, while also achieving competitive results on two publicly available standard datasets, as SemanticKITTI and PandaSet.
The code of our method is available at \href{https://github.com/SiMoM0/3PNet}{https://github.com/SiMoM0/3PNet}.
\end{abstract}




\end{frontmatter}



\section{Introduction}
\label{sec1}

LiDAR semantic segmentation is a fundamental task in 3D scene understanding, essential for applications such as autonomous driving, robotics, and decision-making \citep{guo2020deep}. It involves assigning semantic labels to each point in a LiDAR point cloud, enabling accurate perception of the surrounding environment.

Many state-of-the-art methods \citep{xu2021rpvnet, yan20222dpass, liu2023uniseg} boost performance by combining multiple point cloud representations or incorporating external data sources, such as images.
However, this often increases system complexity and introduces dependencies on additional sensors. Furthermore, these approaches typically require large amounts of data, frequently supplemented by external datasets \citep{liu2023uniseg, wu2024point}, making them less suitable for data-constrained environments.
Although data augmentation is commonly used to improve model generalization, 
many existing techniques do not consider the specific geometric and spatial properties of point clouds acquired with LiDAR sensors, thus limiting their effectiveness in the real world.
\begin{figure}[ht]
    \centering
    \includegraphics[width=0.85\columnwidth]{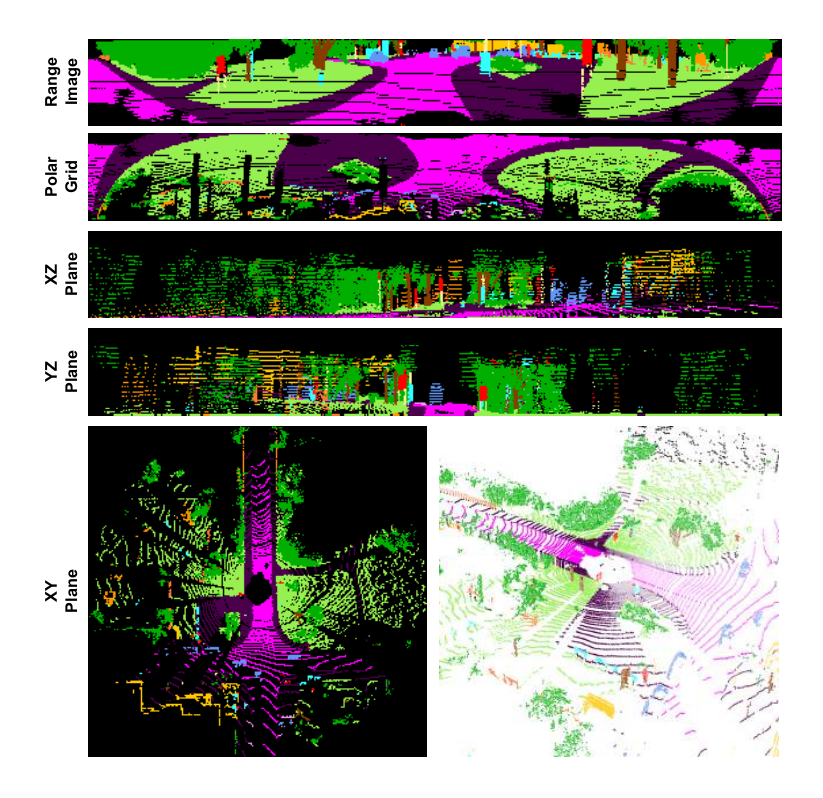}
    \caption{The five 2D point cloud representations obtained through point-plane projections used in our 3D LiDAR Semantic Segmentation method. From top to bottom Range image, Polar grid, XZ plane, YZ plane and XY plane (Bird's-Eye View). Bottom-right: example 3D point cloud from which the 2d representations are extracted.}
    \label{fig:projections}
\end{figure}

Recent LiDAR semantic segmentation methods either operate directly on the raw point-based data or adopt alternative representations such as range-view projections, voxels, or multi-modal, fusion-based data, each offering distinct advantages and limitations.
Range-based methods \citep{wu2018squeezeseg, milioto2019rangenet++, kong2023rethinking} take advantage of 2D representations, enabling fast computations and the integration of 2D CNNs. However, they face challenges related to information loss due to the projection process. In contrast, voxel-based \citep{zhu2021cylindrical, lai2023spherical} and fusion-based approaches \citep{yan20222dpass, liu2023uniseg} leverage rich 3D geometric information and enhance features by incorporating data from multiple sensor perspectives. 
Although these methods often produce better results, they come with trade-offs, such as slower inference times, higher memory consumption, and, in the case of fusion-based approaches, dependence on data from multiple sensors, thus limiting their generalizability to changing light and weather conditions.
Most of these methods also rely on a large amount of annotated data covering most of the domains of interest, thus limiting their adaptability to general applications in real-world scenarios, where often the available data is limited and with a non-negligible domain shift. 
Recent works on self-supervised learning for LiDAR point clouds \citep{nunes2022segcontrast, nisar2025psa} optimizes models with a small percentage of the annotated training data (typically 0.1\% to 10\%) but the split is performed over \emph{all} training sequences, so the data still covers all environments represented in the dataset.
In this work, we tackle a similar yet more challenging problem: we also focus on small data scenarios, but the training data is extracted from a single sequence, thus not covering the full diversity of the dataset domain.\\

In this paper, we propose \textbf{3PNet}, a novel point-based LiDAR semantic segmentation method for small data scenarios, that exploits \textbf{P}oint-\textbf{P}lane \textbf{P}rojections.
To efficiently leverage local features and geometric properties of the point cloud,  while avoiding the high computational cost of 3D convolutions, we exploit 2D representations (\figref{fig:projections}) as rich source of information and complementary perspectives.
This enables the use of efficient 2D convolution to enhance results and improve performance, combined with simple per-point linear affine layers for refinement at the 3D level.
Additionally, we propose a geometry-aware Instance CutMix augmentation, which adapts to the spatial and geometric properties of LiDAR-generated point cloud, effectively addressing class imbalance. These modifications allow our model to achieve strong performance in data-limited scenarios while maintaining competitive results in standard setups, where the full training data are available.

The key contributions of this work are as follows:

\begin{itemize}
    \item We propose a deep learning architecture that directly processes the raw point cloud and exploits 2D representations to enhance the quality of 3D semantic segmentation.
    \item We propose a geometry-aware Instance CutMix augmentation that aligns with the LiDAR beam configuration, improving results and addressing class imbalance.
    \item We conduct extensive results on two different datasets with diverse LiDAR sensor configurations to test the  robustness and adaptability of our method.
    \item We demonstrate that our method significantly improves performance in scenarios characterized by limited data availability (referred to as "small data" setup).
    \item We release an open-source implementation of the proposed approach.
\end{itemize}

\section{Related Works}

Point cloud semantic segmentation approaches can be grouped into four categories: \textit{point-based}, \textit{projection-based}, \textit{discretization-based} and \textit{hybrid methods}.

\subsection{Point-based Methods}

These methods directly work on the irregular and unordered 3D points, preserving the overall structure. Pioneering work PointNet \citep{qi2017pointnet} and its improved version PointNet++ \citep{qi2017pointnet++} learn per-point features using shared Multi-Layer Perceptrons (MLPs) and symmetrical pooling functions. Several approaches as KPConv \citep{thomas2019kpconv}, DGCNN \citep{wang2019dynamic}, and PointConv \citep{wu2019pointconv} propose point-convolution operator for point clouds to leverage local geometric features. Some methods focus on optimizing point sampling, such as RandLA-Net \citep{hu2020randla}, which incorporates an attention mechanism, and PointASNL \citep{yan2020pointasnl}, designed to effectively handle noisy inputs. Others, like WaffleIron \citep{puy2023using} and WaffleAndRange \citep{fusaro2024exploiting} utilize a 3D backbone that integrates MLPs and 2D convolutions, leveraging two-dimensional planar projections of the point cloud. PointTransformer \citep{zhao2021point, wu2022point, wu2024point} directly process the point cloud using the self-attention mechanism \citep{vaswani2017attention}.

\subsection{Projection-based Methods}

These approaches project the point cloud onto a 2D surface as a pre-processing step, enabling the use of well-established 2D convolutional methods and allowing for faster computation, particularly with LiDAR data. Using a spherical range projection, SqueezeSeg \citep{wu2018squeezeseg} and its improved versions \citep{wu2019squeezesegv2, xu2020squeezesegv3} adopt an encoder-decoder network based on SqueezeNet \citep{iandola2016squeezenet} and refine the predictions with conditional random fields. Top-view images of the point cloud are used in \citep{caltagirone2017fast} while PolarNet \citep{zhang2020polarnet} exploits BEV images on a polar coordinate system. SalsaNet \citep{aksoy2020salsanet} and the improved version SalsaNext \citep{cortinhal2020salsanext} leverage BEV images and range-view images respectively using a ResNet \citep{he2016deep} backbone. RangeNet++ \citep{milioto2019rangenet++} adopts DarkNet to process range images and proposes an efficient GPU-based k-NN post-processing technique. Lite-HDSeg \citep{razani2021lite} introduces a boundary loss to improve the segmentation results while KPRNet \citep{kochanov2020kprnet} integrates KPConv as a final layer. MINet \citep{li2021multi} proposes a multi-scale approach, while \citep{alonso20203d} uses an efficient neighbor-searching strategy prior to projection. FIDNet \citep{zhao2021fidnet} and CENet \citep{cheng2022cenet} reintroduce ResNet as backbone, implementing basic interpolation in the decoder. RangeViT \citep{ando2023rangevit} and RangeFormer \citep{kong2023rethinking} exploit the attention mechanism, adopting vision transformers \citep{dosovitskiy2020image} and pyramid vision transformers \citep{wang2021pyramid} respectively to leverage range-view representations.

\subsection{Discretization-based Methods}

Discretization-based approaches usually represent the point cloud as a 3D grid of voxel and apply 3D convolution operators. SEGCloud \citep{tchapmi2017segcloud} introduces a 3D fully convolutional network to process the voxels, while MinkowskiNet \citep{choy20194d} uses sparse convolution to deal with point cloud sparsity and achieve faster computation. {(AF)$^2$-S3Net} \citep{cheng2021af2s3net} exploits the attention mechanism to aggregate multi-scale features. Cylinder3D \citep{zhu2021cylindrical} partitions the points in a 3D cylindrical grid and proposes asymmetrical convolution to process the voxels. Recent work SphereFormer \citep{lai2023spherical} considers the point distribution, designing a radial window self-attention network, while RetSeg3D \citep{erabati2025retseg3d} employs the retention mechanism \citep{sun2023retentive} to model long-range dependencies.

\subsection{Hybrid Methods}

These methods exploit complementary attributes provided by representations such as points, projection planes, and voxels or introduce data from other sensors such as cameras to improve the segmentation results. SPVCNN \citep{tang2020searching} combines features at point and voxel level using sparse convolution. RPVNet \citep{xu2021rpvnet} proposes a range-point-voxel fusion network to exploit the three different views, while PVKD \citep{hou2022point} introduces a point-to-voxel knowledge distillation module. 2DPASS \citep{yan20222dpass} uses RGB features from camera images to enhance the training process. UniSeg \citep{liu2023uniseg} combines all the three views and RGB images to improve the segmentation results while FRNet \citep{xu2023frnet} proposes a range-point network for improved and fast inference. Recent works such as RAPiD \citep{li2025rapid} focus on learning robust embeddings taking into account point distribution and density issues.\\

This paper is an extension of our previous work \citep{fusaro2024exploiting}, which was inspired by \citep{puy2023using}.  
Compared to these works, 3PNet introduces two novel 2D point cloud representations, targeted architectural improvements that involve the SpatialMix module and the cascade of projections, a geometry-aware Instance CutMix augmentation strategy, and a more comprehensive experimental evaluation demonstrating improved performance.

\section{Method}

In this section, we present the architecture of 3PNet, which consists of three main modules: the point cloud embedding, the backbone utilizing point-plane projections, and the segmentation head. Our approach achieves high-quality semantic segmentation by integrating the most informative 2D plane projections.

\subsection{Point Cloud Embedding}

\begin{figure*}[t]
    \includegraphics[width=\textwidth]{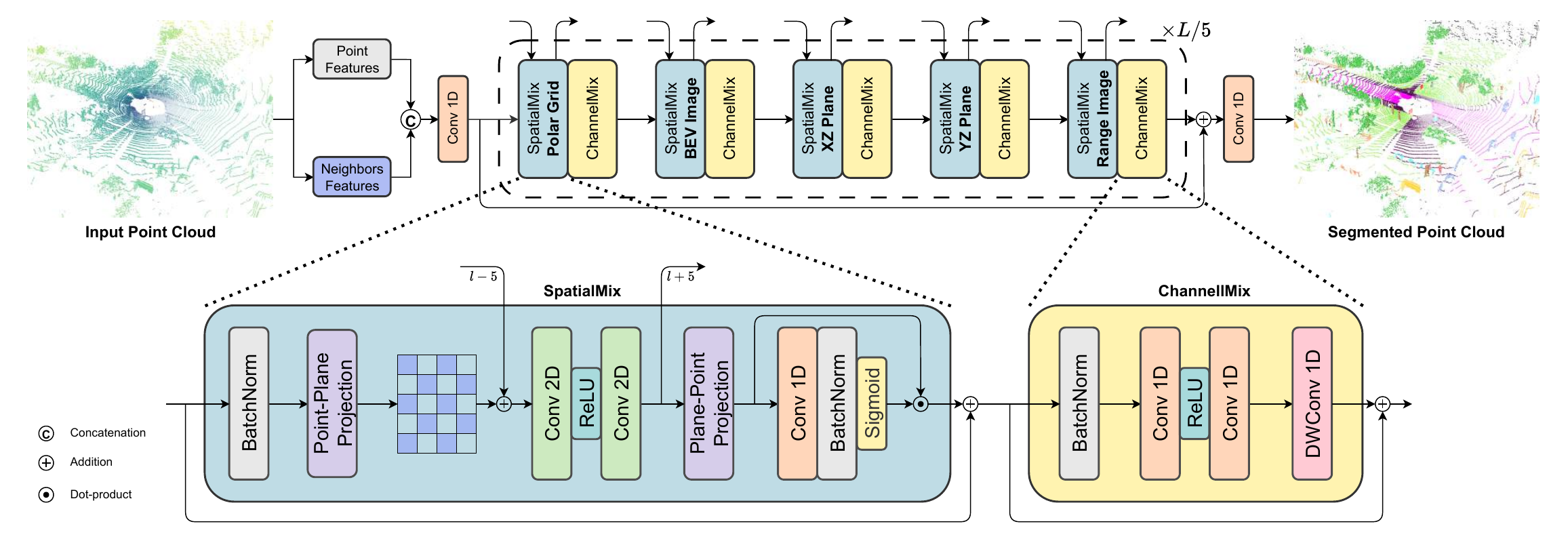}
    \caption{Overall architecture of our proposed method 3PNet. The Point Embedding Module efficiently learns initial per-point features combining information from each point and its corresponding neighborhood. The Backbone incorporates point-plane projections across five different 2D representations, combining 2D convolutions and simple MLPs to produce good features. Finally, the Semantic Head produces the predictions through a final convolution layer.}
    \label{fig:network}
\end{figure*}

Similar to \citep{puy2023using, fusaro2024exploiting}, the entry point of our architecture is represented by a point cloud embedding module that operates directly on raw 3D points to extract initial feature embeddings for each point, combining both per-point and neighboring points features.
3PNet takes as input a point cloud $\mathbf{F_{in}} \in \mathbb{R}^{N \times C_{in}}$ containing $N$ points, each with $C_{in}=5$ features representing the coordinates $x$, $y$, $z$, the remission value  and the range, i.e., the Euclidean distance from the sensor. We define the coordinates of a point as $\mathbf{p}_i = (x,y,z)$ and its associated features as $\mathbf{f}_i \in \mathbb{R}^5$.
Two separate branches process the input point cloud to extract features for individual points and their neighborhoods, respectively. 
The first branch applies to $\mathbf{F_{in}}$ a per-point linear affine layer $\text{LN}$ to map the $C_{in}$-dimensional input features to $C$-dimensional features $\mathbf{f}'_i$:
\begin{equation}
    \mathbf{f}'_i = \text{LN}(\text{BN}(\mathbf{f}_i))
    \label{eq:}
\end{equation}
where $\text{BN}$ denotes the batch normalization operator.
We implemented this mapping efficiently using a simple 1D convolutional layer with a $1 \times 1$ kernel\footnote{The Conv1D blocks in \figref{fig:network} always refer to the efficient implementation of a per-point linear affine layer.} applied to $\mathbf{F_{in}}$.
The second branch employs a KDTree to efficiently extract the K-nearest neighbors for each 3D point $\mathbf{p}_i$, forming a neighborhood of input features $\mathcal{N}_i = \{ \mathbf{f}_{i,1}, \dots, \mathbf{f}_{i,K} \}$. 
We map $\mathcal{N}_i$ into a $C-$dimensional features as:
\begin{equation}
\mathbf{f}''_i = \text{max}_\text{c}\{\text{MLP}(\mathbf{f}_{i,j} - \mathbf{f}_i)\}_{j=1}^{K}
\end{equation}
where MLP denotes a two-layer Multi-Layer Perceptron with input dimension $5$, hidden layer of $256$ units, and output dimension $C$, with a rectified linear activation function between the two layers, and $\text{max}_\text{c}$ is a max-pool operator over channels dimension. Batch normalization is applied to both the input and output features.
By packing all the neighborhood input features  $\mathcal{N}_i$ into a data structure $\mathbf{P_{in,K}} \in \mathbb{R}^{K \times N \times C_i}$, we efficiently implemented the feature mapping by combining 2D convolutions with $1 \times 1$ kernels, batch normalizations, rectified linear units, and a final max pooling layer.
The outputs of the two branches are then concatenated and mapped back to a $C$-dimensional feature for each input point:
\begin{equation}
\mathbf{f}^{(0)}_i = \text{LN}(\mathbf{f}'_i \Vert \mathbf{f}''_i)
\end{equation}
where $\Vert$ denotes concatenation operator.
The resulting $N$ features are stacked into a data structure $\mathbf{P^{(0)}} \in \mathbb{R}^{N \times C}$, which represents the input for the subsequent layers.

\subsection{Point-Plane Projections}


The backbone of our method is composed of a sequence of $L$ layers, each comprising a cascade of two modules: the SpatialMix module, which leverages point-plane projections, and the ChannelMix module, which refines the features at a 3D level. The goal is to extract rich per-point features that capture both local and global relationships among points, enabling accurate LiDAR semantic segmentation.
\subsubsection{SpatialMix Module}
Introduced in \citep{puy2023using}, point-plane projections allow to obtain 2D representations of the point cloud that are used to efficiently extract spatial and geometric information with efficient 2D convolutions, avoiding the need for complex and slow 3D convolution operations. 
The SpatialMix module is designed to efficiently leverage 2D representations of the point cloud and enhance feature learning. 
At each of the $L$ layers, the Spatial Mix module cyclically leverages a different point-plane projection of the point cloud, selecting, in turn, one of the following projection planes: i)  Polar Grid; ii) XY Plane as BEV Image; iii) XZ Plane; iv) YZ Plane; v) Range Image.
Each representation enables to capture a unique perspective of the 3D environment, allowing the network to leverage the complementary strengths of each plane to extract good features. 
The Polar Grid and the Range Image representations are novel additions compared to \citep{puy2023using}, and they provide a noticeable boost in performance (see \secref{sec:ablation}).
Each of these five point-plane projections are repeated $L/5$ times to match the total number of layers $L$, as shown in \figref{fig:network}.\\
At layer $l$, each SpatialMix module takes as input the features processed in the previous layer $\mathbf{F}^{(l-1)}$.
Consider the following recursive layers:
\begin{equation}
\mathbf{P}^{(l)} = 
\begin{cases}
\text{Conv}(\Pi_p(\text{BN}(\mathbf{F}^{(l-1)})))&\quad\text{if}~l\leq5\\
\text{Conv}(\Pi_p(\text{BN}(\mathbf{F}^{(l-1)})) + \mathbf{P}^{(l-5)})&\quad\text{if}~l>5
\end{cases}
\end{equation}
where $\Pi_p$ denotes a point-plane projection function with $p = 1,\dots,5$  the plane index and $\text{Conv}$ denotes a two-layer 2D convolutional operation with $3 \times 3$ kernels, a hidden layer of 256 channels, rectified linear activation function and an output dimension of $C$.
For the XY, XZ, or YZ planes, the function $\Pi_p$ projects in the selected plane all the original points $\mathbf{p}_i$, discretizes the plane into a grid of $w \times h$ cells and, for each cell, averages the current features of all points that fall into that cell. For the Polar Grid, $\Pi_p$ converts each point $\mathbf{p}_i$ into polar coordinates $(\rho, \phi)$ on the horizontal plane, where $\rho = \sqrt{x^2 + y^2}$ and $\phi = \arctan(y, x)$. The space is then discretized into $R$ rings and $S$ sectors, assigning to each cell the average features of the points within it. For each cell $(u,v)$ per-point features are assigned, applying a logarithmic scale along the radial dimension to better match the spatial properties, as follows:
\begin{equation}\label{eq:polar-projection}
    \binom{u}{v} = \binom{\frac{\log(\rho) - \log(\rho_{\min})}{\log(\rho_{\max}) - \log(\rho_{\min})} \cdot (R - 1)}{\frac{1}{2}(\phi + \pi) S \pi^{-1} - 1}
\end{equation}
The grid is then unrolled into a 2D image of size $R \times S$, where height and width correspond to the number of rings and sectors, respectively.
For the Range Image, $\Pi_p$ performs a spherical projection of each point $\mathbf{p}_i = (x,y,z)$ onto a 2D image of size $H \times W$, based on its azimuth and elevation angles. Each pixel $(u,v)$ of the resulting image will contain the correspondent features of one or more points as follows:
\begin{equation}\label{eq:spherical-projection}
	\binom{u}{v} = \binom{\frac{1}{2}[1 - \arctan(y, x)\pi^{-1}]W}{[1 - (\arcsin(z, r^{-1}) + f^{down})f^{-1}]H}
\end{equation}
where $r = \sqrt{x^2 + y^2 + z^2}$ is the range of each point and $f = f^{up} - f^{down}$ is the vertical field-of-view of the sensor. As done for the previous planes, features of all points that fall into a cell are averaged.
For $l \leq 5$, the input to the $\text{Conv}$ operation is just the current projected features, while for $l > 5$, the input to the $\text{Conv}$ operation is the sum of the current projected features and the output of the previous $\text{Conv}$ operation for the same plane type $\mathbf{P}^{(l-5)}$. This simple but powerful layer skip connection preserves important features, improves gradient flow and retains fine-grained details. 
This lightweight 2D convolutional layer enables the model to capture local spatial relationships in a compact and computationally efficient way.
After processing in the 2D space, the obtained features are back-projected to the original 3D points:
\begin{equation}
\mathbf{\hat{F}}^{(l)} = \Pi_p^{-1}(\mathbf{P}^{(l)})
\end{equation}
For each 3D point $\mathbf{p}_i$, $\Pi_p^{-1}$ looks for the 2D cell in which this point falls and copies the corresponding feature from $\mathbf{P}^{(l)}$. The features are then refined by a swish-gated module that selectively enhances relevant information while suppressing noise. It computes a learned weight mask through a 
1D convolutional layer with a $1 \times 1$ kernel followed by a sigmoid activation $\text{Sigmoid}$, generating per-channel importance weights, and combining them with the features through an element-wise multiplication. Finally, to preserve the per-point features\footnote{Otherwise all points projected into the same cell would share the same feature.} and facilitating the gradient flow, a skip connection is applied, combining the input features with the refined ones. The output of the SpatialMix module is given by:
\begin{equation}
    \mathbf{\bar{F}}^{(l)}  = (\mathbf{\hat{F}}^{(l)} \odot \text{Sigmoid}(\text{BN}(\text{Conv1D}(\mathbf{\hat{F}}^{(l)} )))) + \mathbf{F}^{(l-1)}
\end{equation}
where $\odot$ is the dot product operator.\\

By employing different point-plane projections, the model learns robust features from different perspectives, while maintaining low network complexity and ensuring fast and simple processing. 
\subsubsection{ChannelMix Module}
The ChannelMix module is responsible for refining per-point features at the 3D level, enhancing the representation learned from the SpatialMix module. It directly processes in the point cloud space, performing a series of lightweight operations, keeping computational efficiency. ChannelMix employs a batch normalization, followed by two 1D convolutional layers, separated by a ReLU activation function, mirroring the convolutional block in the SpatialMix module. Then it applies a depth-wise 1D convolution, instead of the swish-gated module, followed by a skip connection that combines the input and refined features to preserve original point-wise characteristics:
\begin{equation}
    \mathbf{F}^{(l)}  
    = 
    \text{DWConv1D}(\text{Conv1D}(\text{ReLU}(\text{Conv1D}(\text{BN}(\mathbf{\bar{F}}^{(l)})))))
\end{equation}
where $\text{ReLU}$ denoted the rectified linear activation function and $\text{DWConv1D}$ denotes the depth-wise 1D convolution.\\

The SpatialMix and ChannelMix modules are applied sequentially $L$ times, varying in turn the projection plane for the former.



\subsection{Segmentation Head}

The final module of the network is responsible for producing the final predictions. It first combines the output features from the backbone $\mathbf{F}^{(L)}$ with $\mathbf{P^{(0)}}$, the initial embeddings of each point neighborhood, through a simple addition, acting as a skip connection. This facilitates the direct flow of important geometrical features, preserving spatial details and improving overall performance. Next, a 1D convolutional layer maps the combined features to the final representation, followed by a softmax operator to produce the network predictions.

\subsection{Geometry-Aware Instance CutMix Augmentation}

\begin{figure}[ht]
    \centering
    \includegraphics[width=0.7\linewidth]{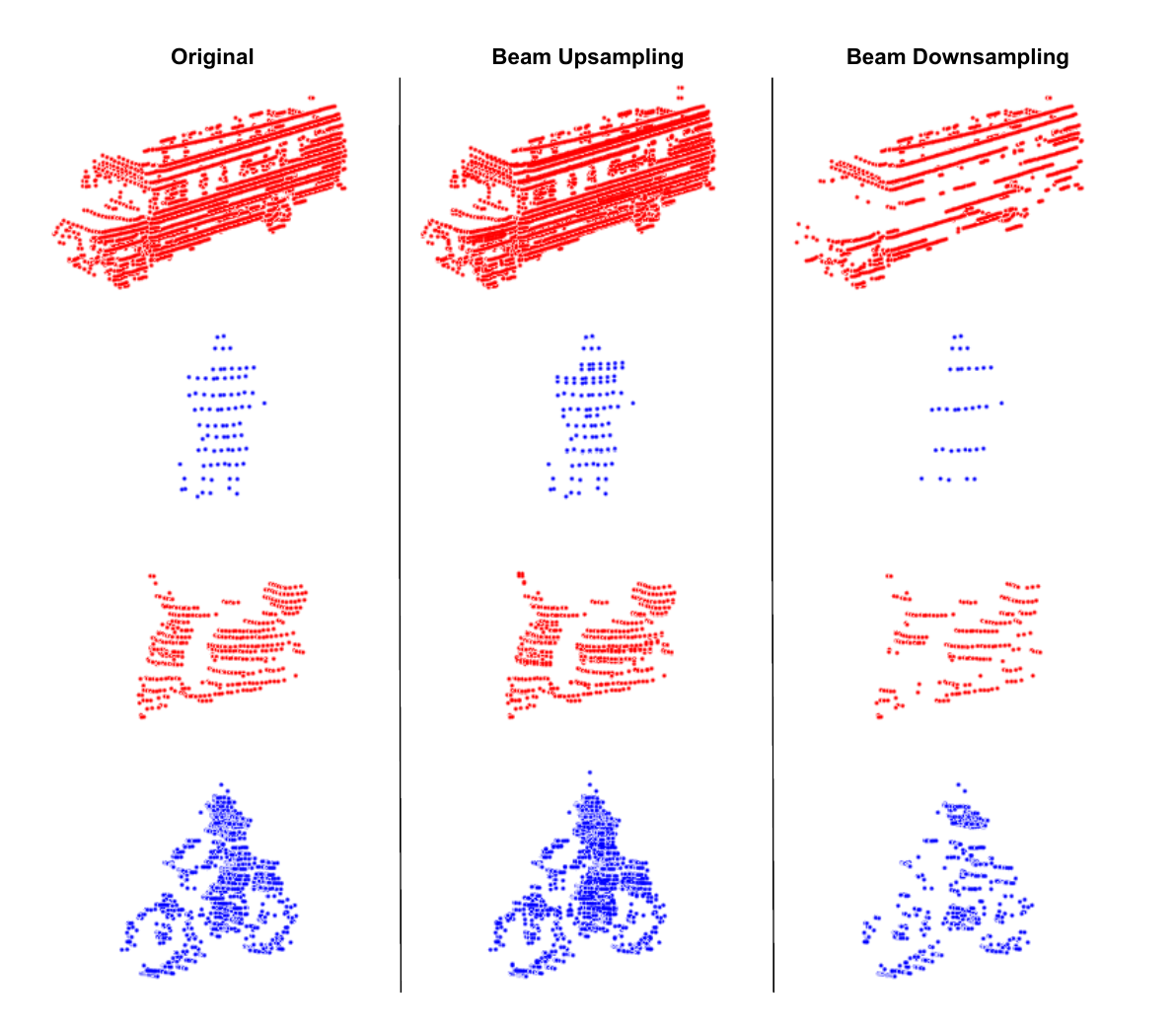}
    \caption{Examples of the proposed Geometry-Aware Instance CutMix augmentation. From top to bottom, instances of \textit{other-vehicle}, \textit{person}, \textit{motorcycle} and \textit{bicyclist} are shown, along with the LiDAR beam-aligned resampling.}
    \label{fig:cutmix-example}
\end{figure}

Following \citet{xu2021rpvnet}, we employ Instance CutMix during training, an augmentation technique specifically designed to mitigate class imbalance in LiDAR semantic segmentation.
It extracts instances of underrepresented classes from the training set and pastes them into new scenes to improve network generalization. To further refine this augmentation, we introduce an adaptive geometry-aware resampling strategy that adjusts the density of each pasted instance based on its placement.
Specifically, it quantizes instances along the $z$-axis, in order to sample beam-like groups of points. In case of upsampling, selected groups are duplicated with a small vertical shift to simulate new LiDAR beams, while for downsampling, groups of points are removed, as shown in \figref{fig:cutmix-example}.
Instances inserted closer to the sensor are upsampled, while those farther away are downsampled, following a linear scale. This adjustment ensures consistency with the LiDAR sensor beams configuration, maintaining a realistic spatial point distribution and improving the segmentation results on underrepresented classes with respect to the standard Instance CutMix augmentation (see \secref{sec:ablation}).

\subsection{Loss Function}

We adopt a loss function that is the weighted sum of the standard cross-entropy loss and the Lovasz-Softmax loss \citep{berman2018lovasz}, which directly optimizes the intersection-over-union ($IoU$) metric. The final loss is defined as follow:

\begin{equation}
    \mathcal{L} = \mathcal{L}_{CE} + \lambda\mathcal{L}_{Lovasz}
\end{equation}

where $\mathcal{L}_{CE}$ is the cross-entropy loss, $\mathcal{L}_{Lovasz}$ is the Lovasz loss, and $\lambda=1.0$ is a weighting factor that controls the contribution of the Lovasz term.

\section{Experiments}


The experimental evaluation is designed to support our claims that the proposed approach outperforms existing methods in LiDAR semantic segmentation with limited data, achieves competitive results on standard publicly available datasets, and that the geometry-aware data augmentation improves class imbalance handling across different models.


\subsection{Datasets}\label{sec:datasets}

We evaluate our approach on two autonomous driving datasets - SemanticKITTI and PandaSet - captured using different LiDAR sensors. We ran the experiments using two distinct dataset splits: the conventional split commonly used for performance evaluation, referred to as the \emph{standard setup}, and a split containing a very limited amount of data from a single sequence, which we refer to as the \emph{small data setup}.

SemanticKITTI \citep{semantickittidataset} contains 22 sequences with 19 semantic classes, captured by a Velodyne HDL-64E LiDAR. In the standard setup, we follow the standard dataset split: sequences 00 to 10 (excluding 08) are used for training, sequence 08 for validation, and the remaining 11 sequences are used as the test set. In the  small data setup, following \citet{fusaro2024exploiting}, only sequence 04, the smallest one, containing only 271 LiDAR point clouds is used for training.

PandaSet \citep{xiao2021pandaset} comprises 103 sequences of 8 seconds length, collected by a Hesai-Pandar64 LiDAR, but only 76 have pointwise annotations resulting in 6080 labeled scans. In the standard setup, following \citet{duerr2020lidar}, we group the semantic classes into 14 for training and evaluation, using the same train/validation/test split of 4320/640/1120 scans. In the  small data setup, to adopt a similar setup to SemanticKITTI, sequences 30, 72, and 115 are selected, resulting in a total of 240 scans for training. These sequences are chosen for their relatively simple environment with fewer distinct classes.

\subsection{Implementation details}

We train our model from scratch using the AdamW optimizer with a batch size of 4, for 45 epochs on two NVIDIA A40 GPUs. The learning rate follows a scheduler with linear warm-up from 0 to 0.002 over the first 4 epochs, then gradually decreases following a cosine annealing schedule down to $10^{-5}$.

The input point cloud is first downsampled using a voxelization operation with resolution 0.1\,m, to reduce the number of points and the computational overhead. It is then cropped to the sensor field of view to minimize outliers and exclude points that are too distant from the sensor, which could negatively impact the projection onto 2D planes. For SemanticKITTI, we use a range of $(-50\,m, 50\,m)$ along the $x$ and $y$ axes, $(-3\,m, 2\,m)$ for the $z$-axis. For PandaSet, due to the different sensor configuration, we use a range of $(-100\,m, 100\,m)$ for the $x$, $y$ axes and $(-2\,m, 10\,m)$ along the $z$-axis. During training, both input points and labels are downsampled and cropped to optimize memory and speed. However, during validation and testing, the full original point cloud is recovered: for points that were excluded due to downsampling or cropping, features are assigned using nearest neighbor interpolation from the processed point cloud. This guarantees that to each original point, a prediction is assigned.

In the Point Cloud Embedding module, $K=16$ nearest neighbors are considered for each point. During training, we apply standard data augmentation techniques, including random rotation around the $z$-axis, random flipping along the $x, y$ axes, and random scaling. For SemanticKITTI standard setup, we also adopt our geometry-aware Instance CutMix, combined with PolarMix \citep{xiao2022polarmix} to better handle class imbalance. We use $L=50$ layers and $C=256$ point features for SemanticKITTI and PandaSet.

\subsection{Metrics}

Following the standard practice in Semantic Segmentation, we utilize the mean intersection over-union (mIoU) over all classes, defined as:

\begin{equation}
    mIoU = \frac{1}{C} \sum_{c=1}^{C} \frac{\textit{TP}_c}{\textit{TP}_c + \textit{FP}_c + \textit{FN}_c},
\end{equation}
where $\textit{TP}_c$, $\textit{FP}_c$, $\textit{FN}_c$ are the true positive, false positive and false negative for $c$ class and $C$ is the number of classes.

\subsection{Performance on Small Data Setup}

We evaluate the effectiveness of our proposed approach in limited-data scenarios to assess its generalization capabilities. We trained all tested methods by using the small data setup splits described in \secref{sec:datasets}. We used all the official open-source implementations provided by the authors. For each method, we applied the recommended parameters specific to each dataset, repeated the training five times, and reported the best results in \tabref{tab:val-semantickitti-sdata}.

3PNet achieves a significant improvement (+1.9\,\%) over our previous approach \citep{fusaro2024exploiting} and outperforms other state-of-the-art methods by a good margin. 
This improvement is attributed to the use of multiple point-plane projections connected by layer skip connections, which enhance the learned features by incorporating complementary geometric information. Additionally, our geometry-aware Instance CutMix augmentation further boosts the results, particularly for underrepresented classes such as other-vehicle and person. Notably, the augmentation technique also improves performance on classes like pole, which shares a similar 3D structure with person. By increasing the presence of underrepresented classes, the model learns more distinctive features, enhancing its ability to differentiate between visually similar categories. Moreover, these results further confirm the overall superiority of point-based methods in the case of limited training data.

\begin{table*}[t]
\centering
\caption{Semantic segmentation performance on SemanticKITTI small data setup. Methods were trained only on Sequence 04 and tested on the validation set (Sequence 08). The \xmark~symbol indicates classes not present in the training set. CutMix refers to our Geometry-Aware Instance CutMix augmentation. Regarding 2DPASS*, we report the results of the baseline of \citep{yan20222dpass} trained with LiDAR data but no images. The best results are shown in bold, the second best are underlined.}
\label{tab:val-semantickitti-sdata}
\resizebox{1.0\textwidth}{!}{%
\begin{tabular}{l |c|ccccccccccccccccccc}
\hline 
Method & \rb{mIoU\,\%} & \rb{car} & \rb{bicycle} & \rb{motorcycle} & \rb{truck} & \rb{other-vehicle} &  \rb{person} & \rb{bicyclist} & \rb{motorcyclist} & \rb{road} & \rb{parking} & \rb{sidewalk} & \rb{other-ground} & \rb{building} & \rb{fence} & \rb{vegetation}  & \rb{trunk} & \rb{terrain} & \rb{pole} & \rb{traffic-sign} \\ 
\hline
RangeFormer \citep{kong2023rethinking} & 15.6 & 47.9 & \xmark & \xmark & \xmark & 0.2 & 0.1 & \xmark & \xmark & 62.5 & 0.1 & 27.9 & \underline{1.3} & 26.6 & 1.8 & 57.4 & 2.6 & 41.1 & 15.6 & 11.8 \\
2DPASS* \citep{yan20222dpass} & 18.9 & 43.7 & \xmark & \xmark & \xmark & 0.3 & \underline{7.6} & \xmark & \xmark & 31.4 & 0.4 & 28.2 & 0.2 & 44.4 & 8.4 & 72.7 & 13.5 & 53.1 & 27.0 & 27.3 \\
FRNet \citep{xu2023frnet} & 22.0 & 71.3 & \xmark & \xmark & \xmark & 0.6 & 0.1 & \xmark & \xmark & \underline{65.2} & 0.2 & \underline{47.4} & \textbf{1.9} & 52.3 & 2.0 & 76.8 & 2.2 & 59.5 & 14.5 & 24.3 \\
Cylinder3D \citep{zhu2021cylindrical} & 23.7 & 68.7 & \xmark & \xmark & \xmark & 0.8 & 1.7 & \xmark & \xmark & 64.0 & 0.3 & 44.3 & 1.1 & 57.5 & 8.5 & 78.8 & 16.4 & 57.9 & 24.6 & 25.3 \\
WaffleIron \citep{puy2023using} & 24.6 & \underline{86.8} & \xmark & \xmark & \xmark & 0.1 & 1.5 & \xmark & \xmark & 60.8 & \underline{0.9} & 43.6 & 0.5 & 55.1 & 4.8 & \underline{82.0} & \underline{18.7} & \textbf{67.6} & 25.5 & 19.2 \\ 
WaffleAndRange \citep{fusaro2024exploiting} & 24.9 & 86.4 & \xmark & \xmark & \xmark & 0.2 & 1.2 & \xmark & \xmark & 61.9 & 0.5 & 47.3 & 0.5 & 55.9 & 4.7 & 80.4 & 18.5 & \underline{66.9} & 24.9 & 23.4 \\
\rowcolor{lightgray!30!white} \textbf{3PNet (Our)} & \underline{26.8} & 83.9 & \xmark & \xmark & \xmark & \underline{1.3} & 0.2 & \xmark & \xmark & \textbf{68.3} & 0.5 & 44.5 & 0.7 & \textbf{64.2} & \textbf{12.8} & 81.4 & 18.6 & 66.4 & \underline{33.8} & \textbf{32.7} \\
\hline
\rowcolor{lightgray!30!white} \textbf{3PNet + CutMix (Our)} & \textbf{27.8} & \textbf{87.9} & \xmark & \xmark & \xmark & \textbf{5.5} & \textbf{14.8} & \xmark & \xmark & 63.7 & \textbf{1.6} & \textbf{48.1} & 0.7 & \underline{60.7} & \underline{9.2} & \textbf{82.2} & \textbf{19.5} & 64.8 & \textbf{41.4} & \underline{28.0} \\
\hline
\end{tabular}}
\end{table*}

For PandaSet, we adopt a setup similar to SemanticKITTI, by selecting sequences 30, 72 and 115, resulting in a total of 240 scans for training. These sequences are chosen for their relatively simple environment with fewer distinct classes. Following \citet{duerr2020lidar}, we evaluate our model on both the validation and test sets. To ensure a fair comparison, we also trained state-of-the-art methods on PandaSet, using their publicly available code. As shown in \tabref{tab:pandaset-smalldata}, our proposed approach outperforms previous works on both validation and test sets, demonstrating its adaptability to point clouds captured by different LiDAR sensors. These improvements highlight the effectiveness of our network in learning meaningful features, and enhancing generalization ability in limited-data scenarios.
We do not apply our geometry-aware Instance Cutmix augmentation on PandaSet, as instance-level labels are not available at point level.

\begin{table*}[t]
\centering
\caption{Semantic segmentation performance on PandaSet small data setup. Methods trained only on sequences 30, 72 and 115. The \xmark~symbol indicates classes not present in the training set. Regarding 2DPASS*, we report the results of the baseline of \citep{yan20222dpass} trained with LiDAR data but no images. The best results are shown in bold, the second best are underlined.}
\label{tab:pandaset-smalldata}
\resizebox{1.0\textwidth}{!}{%
\begin{tabular}{cl|c|cccccccccccccc}
\hline
& Method & \rb{mIoU\,\%} & \rb{car} & \rb{bicycle} & \rb{motorcycle} & \rb{truck} & \rb{other-vehicle} & \rb{person} & \rb{road} & \rb{road barriers} & \rb{sidewalk} & \rb{building} & \rb{vegetation} & \rb{terrain} & \rb{background} & \rb{traffic sign} \\ 
\hline
\multirow{6}{*}{\rotatebox{90}{Validation}} & RangeFormer \citep{kong2023rethinking} & 21.7 & 62.1 & \xmark & \xmark & 0.0 & 7.6 & 1.3 & 67.6 & 0.0 & 34.9 & 55.4 & 44.0 & 10.3 & 18.9 & 1.7 \\
& Cylinder3D \citep{zhu2021cylindrical} & 32.5 & 85.3 & \xmark & \xmark & \textbf{0.1} & 6.0 & \underline{40.6} & 77.3 & 0.0 & 45.1 & 74.7 & 69.4 & 19.1 & 34.7 & \underline{2.6} \\
& 2DPASS* \citep{yan20222dpass} & 33.5 & \textbf{87.9} & \xmark & \xmark & 0.0 & \textbf{27.1} & \textbf{50.2} & 76.6 & 0.0 & 48.1 & 74.1 & 68.5 & 15.1 & 31.7 & 0.2 \\
& WaffleIron \citep{puy2023using} & \underline{34.4} & 85.0 & \xmark & \xmark & 0.0 & 17.5 & 38.5 & 78.8 & 0.0 & 44.6 & 77.1 & \textbf{75.5} & \textbf{21.7} & \underline{39.3} & 1.9 \\
& WaffleAndRange \citep{fusaro2024exploiting} & \underline{34.4} & 85.0 & \xmark & \xmark & 0.0 & 21.0 & 29.6 & \underline{79.6} & 0.0 & \underline{48.8} & \textbf{78.6} & \underline{75.3} & \underline{20.5} & \textbf{39.6} & \textbf{3.3} \\
\rowcolor{lightgray!30!white} & \textbf{3PNet (Our)} & \textbf{34.6} & \underline{87.0} & \xmark & \xmark & 0.0 & \underline{23.8} & 31.2 & \textbf{80.8} & 0.0 & \textbf{56.3} & \underline{77.4} & 74.9 & 13.2 & 38.5 & 1.0 \\
\hline
\hline
\multirow{6}{*}{\rotatebox{90}{Test}} & RangeFormer \citep{kong2023rethinking} & 21.7 & 60.8 & \xmark & \xmark & 0.0 & 3.8 & 1.1 & 68.5 & \textbf{0.6} & 32.6 & 43.9 & 48.2 & 12.6 & 23.0 & \textbf{8.9} \\
& Cylinder3D \citep{zhu2021cylindrical} & 34.8 & \underline{86.1} & \xmark & \xmark & 0.0 & 43.3 & \underline{45.4} & 76.1 & 0.3 & 36.9 & 68.4 & 67.3 & 21.6 & 40.3 & 2.2 \\
& 2DPASS* \citep{yan20222dpass} & 35.4 & 85.9 & \xmark & \xmark & 0.0 & \textbf{55.3} & \textbf{48.5} & 77.5 & \underline{0.4} & 40.6 & 64.9 & 64.8 & 17.9 & 37.5 & 1.7 \\
& WaffleIron \citep{puy2023using} & \underline{36.4} & \textbf{86.8} & \xmark & \xmark & 0.0 & \underline{48.7} & 24.8 & \underline{79.8} & 0.0 & 44.4 & \underline{73.8} & \underline{76.3} & \underline{27.5} & \underline{44.8} & 2.5 \\
& WaffleAndRange \citep{fusaro2024exploiting} & \underline{36.4} & 84.8 & \xmark & \xmark & 0.0 & 44.3 & 25.9 & 79.2 & \textbf{0.6} & \underline{45.9} & \textbf{73.9} & \textbf{76.6} & \textbf{28.6} & \textbf{45.0} & \underline{4.8} \\
\rowcolor{lightgray!30!white} & \textbf{3PNet (Our)} & \textbf{36.9} & \underline{86.1} & \xmark & \xmark & 0.0 & 48.0 & 38.3 & \textbf{80.4} & \textbf{0.6} & \textbf{50.3} & 72.2 & 75.7 & 20.8 & 42.3 & 2.5 \\
\hline
\end{tabular}}
\end{table*}

\subsection{Performance on Standard Setup}

To assess the generalization ability gained in the small data setup, we evaluate our method on the standard setup splits, following the standard benchmarks for SemanticKITTI and PandaSet (see \secref{sec:datasets}). 
The results of the compared methods, reported in \tabref{tab:test-semantickitti}, were taken from the respective papers. For SemanticKITTI, we adopt for 3PNet the same training protocol as other approaches, utilizing both the training and validation splits, and applying test-time augmentations during inference. The results on the test set demonstrate that our method outperforms all other point-based methods, and ranks among the top methods overall, despite not relying on external data or additional modalities such as RGB images.

\begin{table*}[ht]
\centering
\caption{Semantic segmentation performance on SemanticKITTI test set. This table contains methods that do not use external data or images for training. Regarding 2DPASS*, we report the results of the baseline of \citep{yan20222dpass} trained with LiDAR data but no images. The best results are shown in bold, the second best are underlined.}
\label{tab:test-semantickitti}
\resizebox{1.0\textwidth}{!}{%
\begin{tabular}{l |c|ccccccccccccccccccc}
\hline 
Method & \rb{mIoU\,\%} & \rb{car} & \rb{bicycle} & \rb{motorcycle} & \rb{truck} & \rb{other-vehicle} & \rb{person} & \rb{bicyclist} & \rb{motorcyclist} & \rb{road} & \rb{parking} & \rb{sidewalk} & \rb{other-ground} & \rb{building} & \rb{fence} & \rb{vegetation}  & \rb{trunk} & \rb{terrain} & \rb{pole} & \rb{traffic-sign} \\
\hline
CENet \citep{cheng2022cenet} & 64.7 & 91.9 & 58.6 & 50.3 & 40.6 & 42.3 & 68.9 & 65.9 & 43.5 & 90.3 & 60.9 & 75.1 & 31.5 & 91.0 & 66.2 & 84.5 & 69.7 & 70.0 & 61.5 & 67.6 \\
SPVNAS \citep{tang2020searching} & 66.4 & 97.3 & 51.5 & 50.8 & \textbf{59.8} & 58.8 & 65.7 & 65.2 & 43.7 & 90.2 & 67.6 & 75.2 & 16.9 & 91.3 & 65.9 & 86.1 & 73.4 & 71.0 & 64.2 & 66.9 \\
2DPASS* \citep{yan20222dpass} & 67.4 & 96.3 & 51.1 & 55.8 & \underline{54.9} & 51.6 & 76.8 & \underline{79.8} & 30.3 & 89.8 & 62.1 & 73.8 & 33.5 & 91.9 & 68.7 & \underline{86.5} & 72.3 & \underline{71.3} & 63.7 & 70.2 \\
(AF)$^2$-S3Net \citep{cheng2021af2s3net} & 69.7 & 94.5 & 65.4 & \textbf{86.8} & 39.2 & 41.1 & \textbf{80.7} & \textbf{80.4} & \textbf{74.3} & 91.3 & 68.8 & 72.5 & \textbf{53.5} & 87.9 & 63.2 & 70.2 & 68.5 & 53.7 & 61.5 & \underline{71.0} \\
RetSeg3D \citep{erabati2025retseg3d} & 70.3 & 97.3 & 67.4 & 69.2 & 52.2 & \underline{63.2} & 77.0 & 76.1 & \underline{54.7} & 90.1 & 68.7 & 74.5 & 21.9 & 92.9 & 69.9 & 84.6 & 72.8 & 68.8 & 64.0 & 69.8 \\
RPVNet \citep{xu2021rpvnet} & 70.3 & \textbf{97.6} & \underline{68.4} & 68.7 & 44.2 & 61.1 & 75.9 & 74.4 & 43.4 & \textbf{93.4} & \underline{70.3} & \textbf{80.7} & 33.3 & \textbf{93.5} & \underline{72.1} & \underline{86.5} & \underline{75.1} & \textbf{71.7} & 64.8 & 61.4 \\
SDSeg3D \citep{li2022self} & 70.4 & \underline{97.4} & 58.7 & 54.2 & \underline{54.9} & \textbf{65.2} & 70.2 & 74.4 & 52.2 & 90.9 & 69.4 & 76.7 & \underline{41.9} & \underline{93.2} & 71.1 & 86.1 & 74.3 & 71.1 & 65.4 & 70.6 \\
WaffleIron \citep{puy2023using} & \underline{70.8} & 97.2 & \textbf{70.0} & \underline{69.8} & 40.4 & 59.6 & \underline{77.1} & 75.5 & 41.5 & 90.6 & \textbf{70.4} & 76.4 & 38.9 & \textbf{93.5} & \textbf{72.3} & \textbf{86.7} & \textbf{75.7} & \textbf{71.7} & \underline{66.2} & \textbf{71.9} \\
\rowcolor{lightgray!30!white} \textbf{3PNet (Our)} & \textbf{70.9} & 97.0 & 68.0 & 67.5 & 50.0 & 60.5 & 76.2 & 72.3 & 43.1 & \underline{92.9} & 69.7 & \underline{79.9} & 37.4 & 92.8 & 71.4 & 85.9 & 74.7 & 70.5 & \textbf{66.5} & \textbf{71.9} \\
\hline
\end{tabular}}
\end{table*}

\tabref{tab:test-pandaset} presents the results on the PandaSet test set, highlighting the superiority of our method among those that rely only on point cloud data, without utilizing camera inputs or memory from previous scans. Furthermore, our approach matches or even outperforms methods that incorporate image data or temporal memory. These results reinforce our claim that integrating point-plane projections across multiple representations within the backbone enhances the network's ability to learn complementary features and adapt to different LiDAR sensor configurations.

\begin{table*}[ht]
\centering
\caption{Semantic segmentation performance on PandaSet test set. The "Modalities" column indicates the input type: L for point cloud, C for RGB images, and M for memory, where previous scans are used to improve the current predictions. $^{\dag}$ indicates results reproduced with the publicly released code. The best results are shown in bold, the second best are underlined.}
\label{tab:test-pandaset}
\resizebox{1.0\textwidth}{!}{%
\begin{tabular}{l|c|c|cccccccccccccc}
\hline 
Method & \rb{modalities} & \rb{mIoU\,\%} & \rb{car} & \rb{bicycle} & \rb{motorcycle} & \rb{truck} & \rb{other-vehicle} & \rb{person} & \rb{road} & \rb{road barriers} & \rb{sidewalk} & \rb{building} & \rb{vegetation} & \rb{terrain} & \rb{background} & \rb{traffic sign} \\ 
\hline
RangeFormer$^{\dag}$ \citep{kong2023rethinking} & L & 47.1 & 86.8 & 17.3 & 14.1 & 13.7 & 26.9 & 47.2 & 86.3 & 14.1 & 62.7 & 79.8 & 76.0 & 55.9 & 53.5 & 24.4 \\
Cylinder3D$^{\dag}$ \citep{zhu2021cylindrical} & L & 61.1 & 95.4 & 26.2 & 30.0 & 36.9 & 70.2 & 74.5 & 89.9 & 19.8 & 68.9 & \underline{88.0} & 86.3 & 61.7 & 66.3 & 29.8 \\
WaffleAndRange$^{\dag}$ \citep{fusaro2024exploiting} & L & 62.8 & 94.7 & 44.4 & 34.2 & 42.0 & 59.5 & \textbf{80.5} & \underline{91.9} & 26.9 & 71.1 & \textbf{88.1} & 88.3 & 63.8 & \underline{68.7} & 24.8 \\
Lidar3DSeg \citep{duerr2021decoupled} & L & 63.1 & \textbf{95.8} & 41.9 & 45.8 & \textbf{55.2} & 61.3 & 65.5 & \textbf{96.2} & 25.4 & \underline{72.5} & 86.8 & \textbf{88.8} & 64.4 & 68.6 & 15.6 \\
WaffleIron$^{\dag}$ \citep{puy2023using} & L & 63.7 & 94.7 & 40.3 & 40.8 & 42.4 & 67.2 & 79.3 & 91.5 & \underline{27.1} & 72.2 & 87.8 & 88.0 & \underline{65.4} & \textbf{68.8} & 25.9 \\
2DPASS$^{\dag}$ \citep{yan20222dpass} & L & 64.2 & \underline{95.7} & \textbf{51.7} & 42.6 & 43.8 & \textbf{74.1} & \underline{80.2} & 90.2 & 15.8 & 67.4 & 87.6 & 87.3 & 60.7 & 68.0 & \underline{33.7} \\
SPVCNN \citep{tang2020searching} & L & \underline{64.7} & \textbf{95.8} & 38.1 & \underline{46.3} & 44.0 & \textbf{74.1} & 78.6 & 91.2 & \textbf{28.3} & 70.3 & 87.2 & 87.5 & 61.5 & 67.5 & \textbf{35.6} \\
\rowcolor{lightgray!30!white} \textbf{3PNet (Our)} & L & \textbf{65.2} & 95.0 & \underline{44.6} & \textbf{46.5} & \underline{49.0} & \underline{71.7} & 78.8 & 90.7 & 23.0 & \textbf{73.1} & 87.5 & \underline{88.7} & \textbf{66.9} & 67.3 & 30.3 \\
\hline
\hline
LaserNet++ \citep{meyer2019sensor} & L+C & 59.7 & 94.4 & 23.6 & 38.5 & 27.3 & 72.1 & 61.2 & 95.2 & 26.6 & 72.3 & 85.5 & 90.6 & 62.7 & 67.7 & 18.1 \\
TemporalLidarSeg \citep{duerr2020lidar} & L+M & 60.0 & 93.7 & 33.6 & 38.0 & 37.1 & 59.9 & \underline{72.0} & 91.1 & 14.6 & 70.6 & \underline{88.2} & 88.4 & 63.8 & 68.4 & 20.7 \\
Fusion3DSeg \citep{duerr2021decoupled} & L+C & 65.2 & 96.4 & \underline{46.8} & \underline{49.1} & 52.9 & 71.3 & 65.6 & \textbf{97.1} & \underline{30.5} & \underline{73.0} & 87.2 & \textbf{90.8} & \underline{64.3} & 69.0 & 19.4 \\
PyFu \citep{schieber2022deep} & L+C & \underline{67.8} & \underline{96.5} & 39.6 & 49.0 & \textbf{75.3} & \textbf{82.2} & 67.2 & \underline{96.3} & \textbf{39.6} & 71.4 & 87.7 & \textbf{90.8} & 60.6 & \underline{70.5} & \underline{22.7} \\
MemorySeg \citep{li2023memoryseg} & L+M & \textbf{70.3} & \textbf{97.2} & \textbf{60.2} & \textbf{58.4} & \underline{62.9} & \underline{74.3} & \textbf{82.6} & 92.1 & 27.7 & \textbf{74.1} & \textbf{89.4} & \underline{90.7} & \textbf{64.9} & \textbf{72.8} & \textbf{36.4} \\
\hline
\end{tabular}}
\end{table*}

\subsection{Ablation study}\label{sec:ablation}

In this section, we present an ablation study to assess the contributions of the modules introduced in our approach.

First, we conduct an ablation study on the architecture of the proposed approach (\tabref{tab:ablation-study}) using the SemanticKITTI small data setup. This analysis demonstrates the effectiveness of the newly introduced modules and modifications to the baseline method. We begin with a baseline model following \citet{fusaro2024exploiting}, systematically optimizing it by determining the optimal order of projection planes, number of layers, and incorporating the full point cloud as input.
From this foundation, we evaluate the improvements introduced in our network. As shown in the table, the addition of the swish-gated module, the Layer Skip and the inclusion of polar grid 2D representation leads to significant performance gains. The introduction of Instance CutMix significantly enhances the results, particularly by improving performance on underrepresented classes. Furthermore, the geometry-aware augmentation further boosts overall accuracy.

\begin{table}[t]
    \centering
    \caption{Ablation study on the proposed segmentation pipeline. Results on SemanticKITTI validation set in small data setup. The best results are shown in bold.}
    \label{tab:ablation-study}
    \resizebox{1.0\textwidth}{!}{ 
    \begin{tabular}{cccccc}
        \hline
        \parbox{3cm}{\centering Swish-Gated \\ Module} & \parbox{2cm}{\centering Layer \\ Skip} & \parbox{2cm}{\centering Polar \\ Grid} & \parbox{2cm}{\centering Instance \\ CutMix} & \parbox{2cm}{\centering Geom. \\ Aware} & mIoU \\
        \hline
        & & & & & 25.8 \\
        \cmark & & & & & 26.0 \\
        & \cmark & & & & 26.1 \\
        \cmark & \cmark & & & & 26.2 \\
        \cmark & \cmark & \cmark & & & 26.8 \\
        \cmark & \cmark & \cmark & \cmark & & 27.6 \\
        \cmark & \cmark & \cmark & \cmark & \cmark & \textbf{27.8} \\
        \hline
    \end{tabular}
    }
\end{table}

The second ablation study focuses on the geometry-aware module for Instance CutMix augmentation (\tabref{tab:augmentation_comparison}), demonstrating its effectiveness in improving performance. We evaluate its impact across point-, range- and voxel-based methods. While range-based methods may be affected by the initial projection step, all other approaches benefit from our geometry-aware module when the full LiDAR point cloud is used as input. Notably, \citep{puy2023using} and \citep{fusaro2024exploiting} apply a cropping strategy during training, selecting a subset of $N$ points within a fixed radius around a chosen center. While this reduces computational cost, it limits the effectiveness of our augmentation, as cropped region may not fully capture the spatial distribution of instances. However, when the complete point cloud is used, aligning instance placement with the LiDAR sensor properties leads to a clear performance improvement.

\begin{table}[ht]
    \centering
    \caption{Ablation study on the geometry-aware module for Instance CutMix augmentation on SemanticKITTI validation set, in small data setup.The best results are shown in bold.}
    \label{tab:augmentation_comparison}
    \resizebox{0.95\textwidth}{!}{ 
    \begin{tabular}{lcccc}
        \hline
        Method & Full PC & Baseline & CutMix & Geom. Aware \\
        \hline
        RangeFormer \citep{kong2023rethinking} & \cmark & 15.6 & \textbf{16.9} & 16.4 \\
        2DPASS \citep{yan20222dpass} & \cmark & 18.9 & 20.1 & \textbf{20.5} \\
        Cylinder3D \citep{zhu2021cylindrical} & \cmark & 24.3 & 25.4 & \textbf{26.0} \\
        WaffleIron \citep{puy2023using} & \xmark & 24.6 & \textbf{25.3} & 25.0 \\
        WaffleIron \citep{puy2023using} & \cmark & 25.3 & 26.3 & \textbf{26.6} \\
        WaffleAndRange \citep{fusaro2024exploiting} & \xmark & 24.9 & \textbf{25.9} & 25.7 \\
        WaffleAndRange \citep{fusaro2024exploiting} & \cmark & 25.5 & 26.6 & \textbf{26.8} \\
        \textbf{3PNet (Our)} & \cmark & 26.5 & 27.6 & \textbf{27.8} \\
        \hline
    \end{tabular}
    }
\end{table}

\subsection{Qualitative results}

\begin{figure}[ht!]
\centering
    \includegraphics[width=\textwidth]{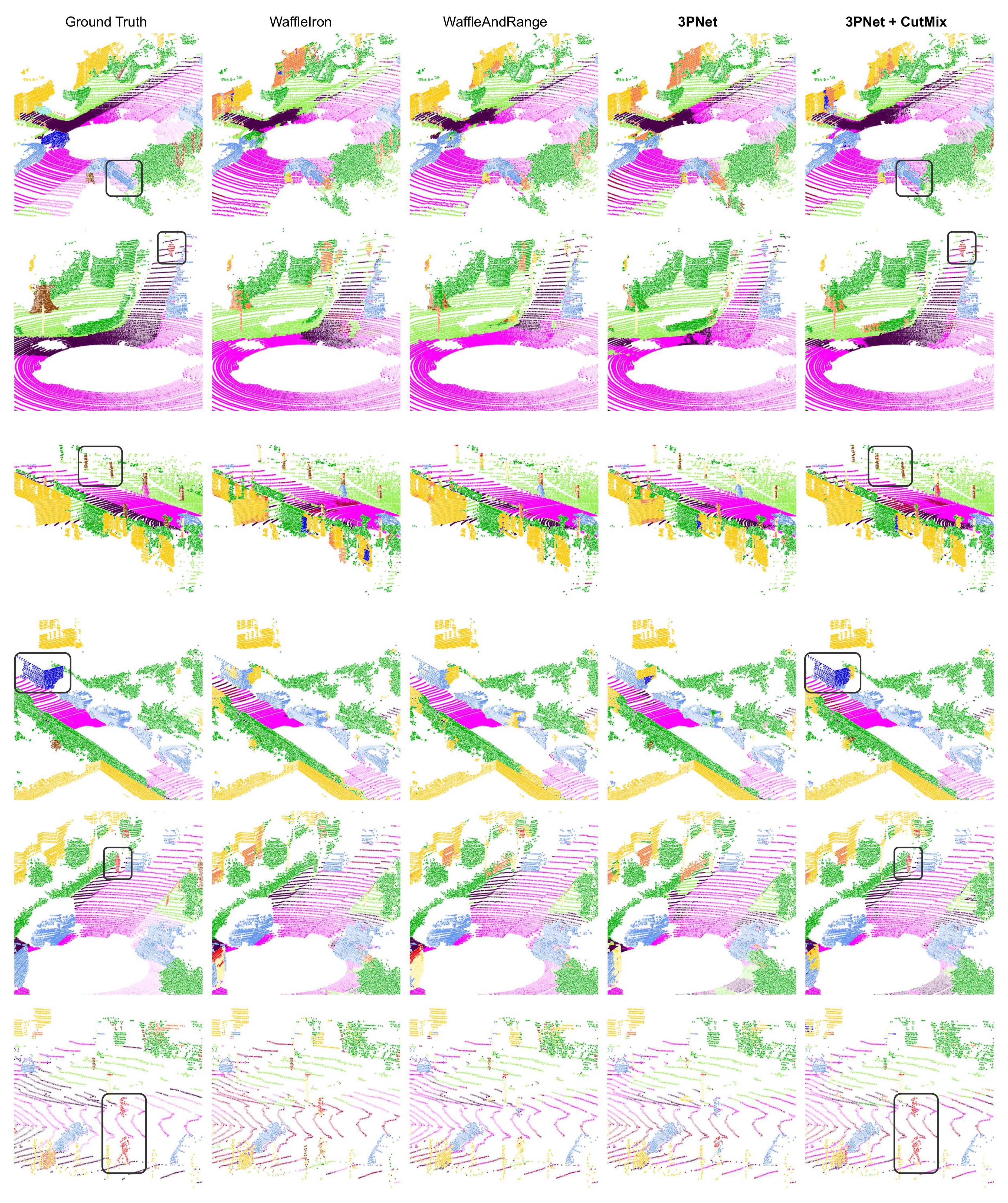}
    \caption{Qualitative comparisons of point-based methods on the SemanticKITTI validation set in small data setup.}
    \label{fig:qualitative_results}
\end{figure}

\begin{figure}[ht!]
\centering
    \includegraphics[width=\textwidth]{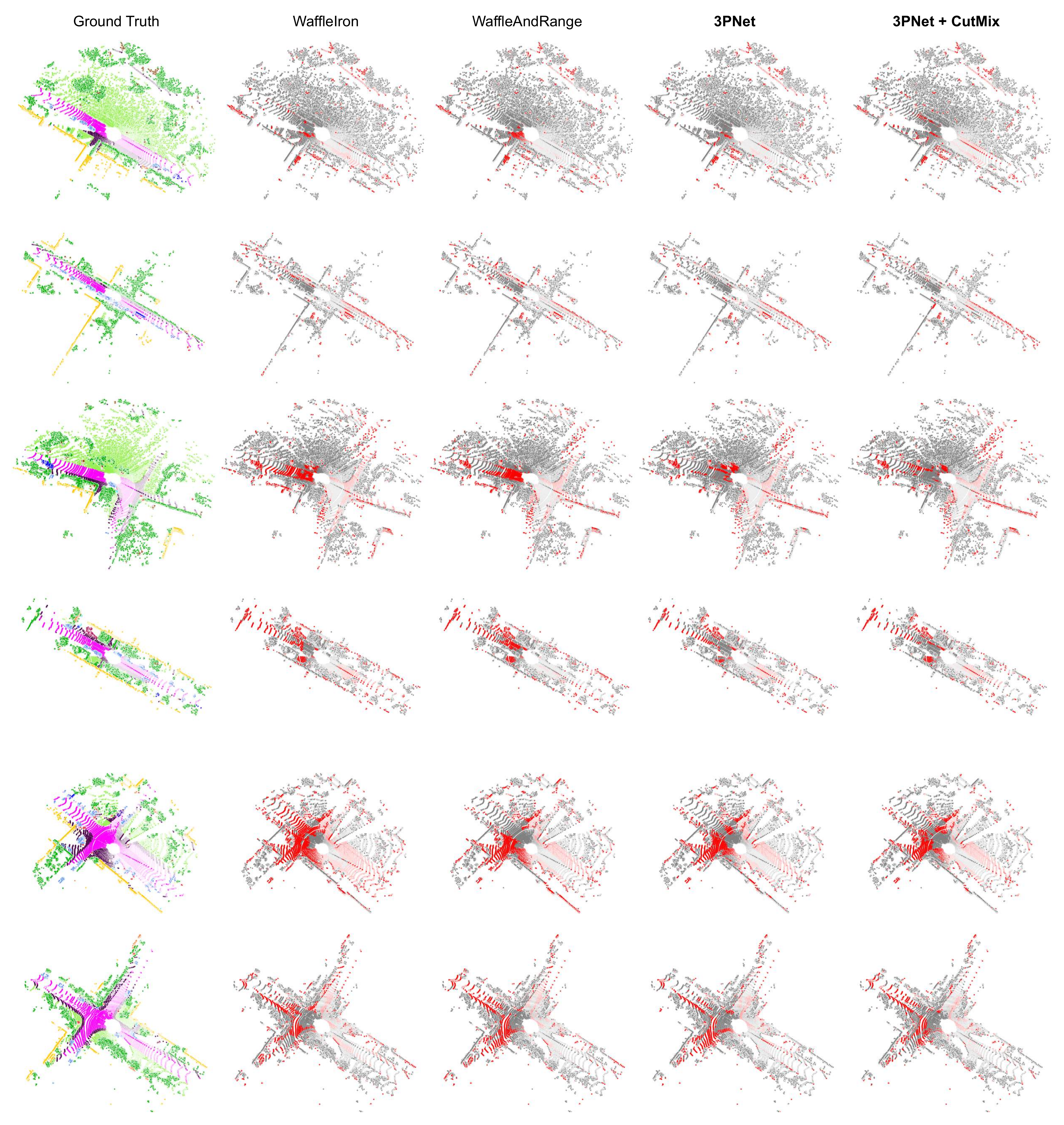}
    \caption{Qualitative comparison of point-based methods on the SemanticKITTI validation set (small data setup), illustrating per-point errors (in red) in the point cloud.}
    \label{fig:error_results}
\end{figure}

\begin{figure}[ht!]
\centering
    \includegraphics[width=0.85\textwidth]{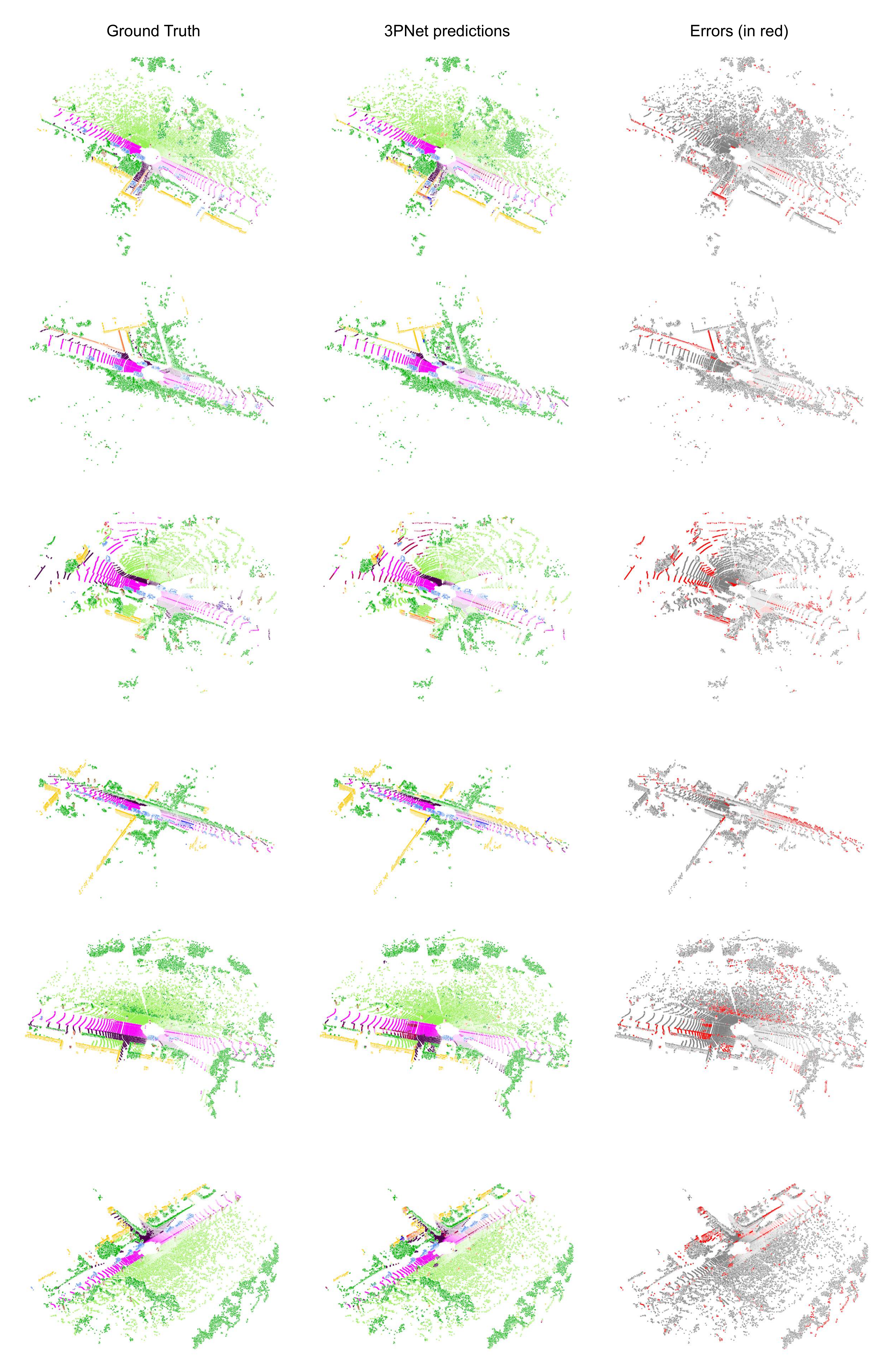}
    \caption{Qualitative results and error visualization of our approach on the SemanticKITTI validation set in small data setup.}
    \label{fig:3pnet_results}
\end{figure}

We present in \figref{fig:qualitative_results} a qualitative comparison between our model, 3PNet, and other point-based methods on SemanticKITTI validation set, in small data setup. Our proposed method demonstrates strong qualitative performance, with significant improvements in underrepresented classes such as person and other-vehicle, as highlighted in the black boxes.
Additional qualitative comparisons are shown in \figref{fig:error_results}, where our model produces significantly fewer errors compared to the other point-based methods.
Furthermore, we provide more qualitative results (\figref{fig:3pnet_results}) of our approach, which clearly show the high accuracy of our predictions, using a small amount of data in training.

\subsection{Inference Time}

We report in \tabref{tab:inference-time} the inference time of our model on SemanticKITTI validation set, using a single NVIDIA A40 GPU. While our model 3PNet results a bit slower, it is important to note that WaffleIron$^{\dag}$ \citep{puy2023using} uses an optimized configuration that merges layers during inference. Our model does not adopt this strategy. Indeed when this technique is not applied, 3PNet achieves a runtime comparable to WaffleIron.

\begin{table}[ht]
    \centering
    \caption{Inference time of different methods estimated on the validation set of SemanticKITTI. $^{\dag}$ denotes model optimization for inference}
    \label{tab:inference-time}
    \resizebox{0.95\textwidth}{!}{ 
    \begin{tabular}{ccccc}
        \hline
        \multirow{2}{*}{Time (ms)} & Cylinder3D & WaffleIron$^{\dag}$ & WaffleIron & 3PNet (Our) \\
         & 139 & 184 & 250 & 234 \\
         \hline
    \end{tabular}}
\end{table}

\section{Conclusions}

In this paper, we introduced a novel approach for LiDAR semantic segmentation that directly leverages 3D point data. Our method employs heterogeneous point-to-plane projections and per-plane skip connections to effectively capture complementary features from multiple perspectives, enabling the use of 2D convolutions for reduced computational complexity and faster inference.
Additionally, we introduced a geometry-aware instance augmentation strategy that enhances performance on underrepresented classes while matching the geometric properties of LiDAR scans. 
We evaluated our approach on multiple datasets and configurations, including sparse data scenarios, where in particular it demonstrated significant improvement over the state of the art. Future work will focus on integrating our architecture into an Unsupervised Domain Adaptation (UDA) framework, enabling effective training using only synthetic data.

\clearpage 


\bibliographystyle{elsarticle-harv} 
\bibliography{glorified, references}

\begin{thebibliography}{60}
\expandafter\ifx\csname natexlab\endcsname\relax\def\natexlab#1{#1}\fi
\providecommand{\url}[1]{\texttt{#1}}
\providecommand{\href}[2]{#2}
\providecommand{\path}[1]{#1}
\providecommand{\DOIprefix}{doi:}
\providecommand{\ArXivprefix}{arXiv:}
\providecommand{\URLprefix}{URL: }
\providecommand{\Pubmedprefix}{pmid:}
\providecommand{\doi}[1]{\href{http://dx.doi.org/#1}{\path{#1}}}
\providecommand{\Pubmed}[1]{\href{pmid:#1}{\path{#1}}}
\providecommand{\bibinfo}[2]{#2}
\ifx\xfnm\relax \def\xfnm[#1]{\unskip,\space#1}\fi
\bibitem[{Aksoy et~al.(2020)Aksoy, Baci and Cavdar}]{aksoy2020salsanet}
\bibinfo{author}{Aksoy, E.E.}, \bibinfo{author}{Baci, S.},
  \bibinfo{author}{Cavdar, S.}, \bibinfo{year}{2020}.
\newblock \bibinfo{title}{Salsanet: Fast road and vehicle segmentation in lidar
  point clouds for autonomous driving}, in: \bibinfo{booktitle}{Proc.~of the
  IEEE Vehicles Symposium (IV)}, pp. \bibinfo{pages}{926--932}.
\bibitem[{Alonso et~al.(2020)Alonso, Riazuelo, Montesano and
  Murillo}]{alonso20203d}
\bibinfo{author}{Alonso, I.}, \bibinfo{author}{Riazuelo, L.},
  \bibinfo{author}{Montesano, L.}, \bibinfo{author}{Murillo, A.C.},
  \bibinfo{year}{2020}.
\newblock \bibinfo{title}{3d-mininet: Learning a 2d representation from point
  clouds for fast and efficient 3d lidar semantic segmentation}.
\newblock \bibinfo{journal}{IEEE Robotics and Automation Letters (RA-L)}
  \bibinfo{volume}{5}, \bibinfo{pages}{5432--5439}.
\bibitem[{Ando et~al.(2023)Ando, Gidaris, Bursuc, Puy, Boulch and
  Marlet}]{ando2023rangevit}
\bibinfo{author}{Ando, A.}, \bibinfo{author}{Gidaris, S.},
  \bibinfo{author}{Bursuc, A.}, \bibinfo{author}{Puy, G.},
  \bibinfo{author}{Boulch, A.}, \bibinfo{author}{Marlet, R.},
  \bibinfo{year}{2023}.
\newblock \bibinfo{title}{Rangevit: Towards vision transformers for 3d semantic
  segmentation in autonomous driving}, in: \bibinfo{booktitle}{Proc.~of the
  IEEE/CVF Conf.~on Computer Vision and Pattern Recognition (CVPR)}, pp.
  \bibinfo{pages}{5240--5250}.
\bibitem[{Behley et~al.(2019)Behley, Garbade, Milioto, Quenzel, Behnke,
  Stachniss and Gall}]{semantickittidataset}
\bibinfo{author}{Behley, J.}, \bibinfo{author}{Garbade, M.},
  \bibinfo{author}{Milioto, A.}, \bibinfo{author}{Quenzel, J.},
  \bibinfo{author}{Behnke, S.}, \bibinfo{author}{Stachniss, C.},
  \bibinfo{author}{Gall, J.}, \bibinfo{year}{2019}.
\newblock \bibinfo{title}{{SemanticKITTI: A Dataset for Semantic Scene
  Understanding of LiDAR Sequences}}, in: \bibinfo{booktitle}{Proc.~of the
  IEEE/CVF Intl.~Conf.~on Computer Vision (ICCV)}, pp.
  \bibinfo{pages}{9297--9307}.
\bibitem[{Berman et~al.(2018)Berman, Triki and Blaschko}]{berman2018lovasz}
\bibinfo{author}{Berman, M.}, \bibinfo{author}{Triki, A.R.},
  \bibinfo{author}{Blaschko, M.B.}, \bibinfo{year}{2018}.
\newblock \bibinfo{title}{The lov{\'a}sz-softmax loss: A tractable surrogate
  for the optimization of the intersection-over-union measure in neural
  networks}, in: \bibinfo{booktitle}{Proc.~of the IEEE/CVF Conf.~on Computer
  Vision and Pattern Recognition (CVPR)}, pp. \bibinfo{pages}{4413--4421}.
\bibitem[{Caltagirone et~al.(2017)Caltagirone, Scheidegger, Svensson and
  Wahde}]{caltagirone2017fast}
\bibinfo{author}{Caltagirone, L.}, \bibinfo{author}{Scheidegger, S.},
  \bibinfo{author}{Svensson, L.}, \bibinfo{author}{Wahde, M.},
  \bibinfo{year}{2017}.
\newblock \bibinfo{title}{Fast lidar-based road detection using fully
  convolutional neural networks}, in: \bibinfo{booktitle}{Proc.~of the IEEE
  Vehicles Symposium (IV)}, pp. \bibinfo{pages}{1019--1024}.
\bibitem[{Cheng et~al.(2022)Cheng, Han and Xiao}]{cheng2022cenet}
\bibinfo{author}{Cheng, H.X.}, \bibinfo{author}{Han, X.F.},
  \bibinfo{author}{Xiao, G.Q.}, \bibinfo{year}{2022}.
\newblock \bibinfo{title}{Cenet: Toward concise and efficient lidar semantic
  segmentation for autonomous driving}, in: \bibinfo{booktitle}{Proc.~of the
  IEEE Intl.~Conf.~on Multimedia and Expo}, pp. \bibinfo{pages}{01--06}.
\bibitem[{Cheng et~al.(2021)Cheng, Razani, Taghavi, Li and
  Liu}]{cheng2021af2s3net}
\bibinfo{author}{Cheng, R.}, \bibinfo{author}{Razani, R.},
  \bibinfo{author}{Taghavi, E.}, \bibinfo{author}{Li, E.},
  \bibinfo{author}{Liu, B.}, \bibinfo{year}{2021}.
\newblock \bibinfo{title}{Af2-s3net: Attentive feature fusion with adaptive
  feature selection for sparse semantic segmentation network}, in:
  \bibinfo{booktitle}{Proc.~of the IEEE/CVF Conf.~on Computer Vision and
  Pattern Recognition (CVPR)}, pp. \bibinfo{pages}{12547--12556}.
\bibitem[{Choy et~al.(2019)Choy, Gwak and Savarese}]{choy20194d}
\bibinfo{author}{Choy, C.}, \bibinfo{author}{Gwak, J.},
  \bibinfo{author}{Savarese, S.}, \bibinfo{year}{2019}.
\newblock \bibinfo{title}{4d spatio-temporal convnets: Minkowski convolutional
  neural networks}, in: \bibinfo{booktitle}{Proc.~of the IEEE/CVF Conf.~on
  Computer Vision and Pattern Recognition (CVPR)}, pp.
  \bibinfo{pages}{3075--3084}.
\bibitem[{Cortinhal et~al.(2020)Cortinhal, Tzelepis and
  Erdal~Aksoy}]{cortinhal2020salsanext}
\bibinfo{author}{Cortinhal, T.}, \bibinfo{author}{Tzelepis, G.},
  \bibinfo{author}{Erdal~Aksoy, E.}, \bibinfo{year}{2020}.
\newblock \bibinfo{title}{Salsanext: Fast, uncertainty-aware semantic
  segmentation of lidar point clouds}, in: \bibinfo{booktitle}{International
  Symposium on Visual Computing (ISVC)}, pp. \bibinfo{pages}{207--222}.
\bibitem[{Dosovitskiy et~al.(2020)Dosovitskiy, Beyer, Kolesnikov, Weissenborn,
  Zhai, Unterthiner, Dehghani, Minderer, Heigold, Gelly
  et~al.}]{dosovitskiy2020image}
\bibinfo{author}{Dosovitskiy, A.}, \bibinfo{author}{Beyer, L.},
  \bibinfo{author}{Kolesnikov, A.}, \bibinfo{author}{Weissenborn, D.},
  \bibinfo{author}{Zhai, X.}, \bibinfo{author}{Unterthiner, T.},
  \bibinfo{author}{Dehghani, M.}, \bibinfo{author}{Minderer, M.},
  \bibinfo{author}{Heigold, G.}, \bibinfo{author}{Gelly, S.}, et~al.,
  \bibinfo{year}{2020}.
\newblock \bibinfo{title}{An image is worth 16x16 words: Transformers for image
  recognition at scale}.
\newblock \bibinfo{journal}{arXiv preprint} .
\bibitem[{Duerr et~al.(2020)Duerr, Pfaller, Weigel and
  Beyerer}]{duerr2020lidar}
\bibinfo{author}{Duerr, F.}, \bibinfo{author}{Pfaller, M.},
  \bibinfo{author}{Weigel, H.}, \bibinfo{author}{Beyerer, J.},
  \bibinfo{year}{2020}.
\newblock \bibinfo{title}{Lidar-based recurrent 3d semantic segmentation with
  temporal memory alignment}, in: \bibinfo{booktitle}{Proc.~of the
  Intl.~Conf.~on 3D Vision (3DV)}, pp. \bibinfo{pages}{781--790}.
\bibitem[{Duerr et~al.(2021)Duerr, Weigel and Beyerer}]{duerr2021decoupled}
\bibinfo{author}{Duerr, F.}, \bibinfo{author}{Weigel, H.},
  \bibinfo{author}{Beyerer, J.}, \bibinfo{year}{2021}.
\newblock \bibinfo{title}{Decoupled iterative deep sensor fusion for 3d
  semantic segmentation}.
\newblock \bibinfo{journal}{International Journal of Semantic Computing} .
\bibitem[{Erabati and Araujo(2025)}]{erabati2025retseg3d}
\bibinfo{author}{Erabati, G.K.}, \bibinfo{author}{Araujo, H.},
  \bibinfo{year}{2025}.
\newblock \bibinfo{title}{Retseg3d: Retention-based 3d semantic segmentation
  for autonomous driving}.
\newblock \bibinfo{journal}{Journal of Computer Vision and Image Understanding
  (CVIU)} \bibinfo{volume}{250}, \bibinfo{pages}{104231}.
\bibitem[{Fusaro et~al.(2024)Fusaro, Mosco, Menegatti and
  Pretto}]{fusaro2024exploiting}
\bibinfo{author}{Fusaro, D.}, \bibinfo{author}{Mosco, S.},
  \bibinfo{author}{Menegatti, E.}, \bibinfo{author}{Pretto, A.},
  \bibinfo{year}{2024}.
\newblock \bibinfo{title}{Exploiting local features and range images for small
  data real-time point cloud semantic segmentation}, in:
  \bibinfo{booktitle}{Proc.~of the IEEE/RSJ Intl.~Conf.~on Intelligent Robots
  and Systems (IROS)}, pp. \bibinfo{pages}{4980--4987}.
\bibitem[{Guo et~al.(2020)Guo, Wang, Hu, Liu, Liu and Bennamoun}]{guo2020deep}
\bibinfo{author}{Guo, Y.}, \bibinfo{author}{Wang, H.}, \bibinfo{author}{Hu,
  Q.}, \bibinfo{author}{Liu, H.}, \bibinfo{author}{Liu, L.},
  \bibinfo{author}{Bennamoun, M.}, \bibinfo{year}{2020}.
\newblock \bibinfo{title}{Deep learning for 3d point clouds: A survey}.
\newblock \bibinfo{journal}{IEEE Trans.~on Pattern Analysis and Machine
  Intelligence (TPAMI)} \bibinfo{volume}{43}, \bibinfo{pages}{4338--4364}.
\bibitem[{He et~al.(2016)He, Zhang, Ren and Sun}]{he2016deep}
\bibinfo{author}{He, K.}, \bibinfo{author}{Zhang, X.}, \bibinfo{author}{Ren,
  S.}, \bibinfo{author}{Sun, J.}, \bibinfo{year}{2016}.
\newblock \bibinfo{title}{Deep residual learning for image recognition}, in:
  \bibinfo{booktitle}{Proc.~of the IEEE/CVF Conf.~on Computer Vision and
  Pattern Recognition (CVPR)}, pp. \bibinfo{pages}{770--778}.
\bibitem[{Hou et~al.(2022)Hou, Zhu, Ma, Loy and Li}]{hou2022point}
\bibinfo{author}{Hou, Y.}, \bibinfo{author}{Zhu, X.}, \bibinfo{author}{Ma, Y.},
  \bibinfo{author}{Loy, C.C.}, \bibinfo{author}{Li, Y.}, \bibinfo{year}{2022}.
\newblock \bibinfo{title}{Point-to-voxel knowledge distillation for lidar
  semantic segmentation}, in: \bibinfo{booktitle}{Proc.~of the IEEE/CVF
  Conf.~on Computer Vision and Pattern Recognition (CVPR)}, pp.
  \bibinfo{pages}{8479--8488}.
\bibitem[{Hu et~al.(2020)Hu, Yang, Xie, Rosa, Guo, Wang, Trigoni and
  Markham}]{hu2020randla}
\bibinfo{author}{Hu, Q.}, \bibinfo{author}{Yang, B.}, \bibinfo{author}{Xie,
  L.}, \bibinfo{author}{Rosa, S.}, \bibinfo{author}{Guo, Y.},
  \bibinfo{author}{Wang, Z.}, \bibinfo{author}{Trigoni, N.},
  \bibinfo{author}{Markham, A.}, \bibinfo{year}{2020}.
\newblock \bibinfo{title}{Randla-net: Efficient semantic segmentation of
  large-scale point clouds}, in: \bibinfo{booktitle}{Proc.~of the IEEE/CVF
  Conf.~on Computer Vision and Pattern Recognition (CVPR)}, pp.
  \bibinfo{pages}{11108--11117}.
\bibitem[{Iandola et~al.(2016)Iandola, Han, Moskewicz, Ashraf, Dally and
  Keutzer}]{iandola2016squeezenet}
\bibinfo{author}{Iandola, F.N.}, \bibinfo{author}{Han, S.},
  \bibinfo{author}{Moskewicz, M.W.}, \bibinfo{author}{Ashraf, K.},
  \bibinfo{author}{Dally, W.J.}, \bibinfo{author}{Keutzer, K.},
  \bibinfo{year}{2016}.
\newblock \bibinfo{title}{Squeezenet: Alexnet-level accuracy with 50x fewer
  parameters and \textless 0.5 mb model size}.
\newblock \bibinfo{journal}{arXiv preprint} .
\bibitem[{Kochanov et~al.(2020)Kochanov, Nejadasl and
  Booij}]{kochanov2020kprnet}
\bibinfo{author}{Kochanov, D.}, \bibinfo{author}{Nejadasl, F.K.},
  \bibinfo{author}{Booij, O.}, \bibinfo{year}{2020}.
\newblock \bibinfo{title}{Kprnet: Improving projection-based lidar semantic
  segmentation}.
\newblock \bibinfo{journal}{arXiv preprint} .
\bibitem[{Kong et~al.(2023)Kong, Liu, Chen, Ma, Zhu, Li, Hou, Qiao and
  Liu}]{kong2023rethinking}
\bibinfo{author}{Kong, L.}, \bibinfo{author}{Liu, Y.}, \bibinfo{author}{Chen,
  R.}, \bibinfo{author}{Ma, Y.}, \bibinfo{author}{Zhu, X.},
  \bibinfo{author}{Li, Y.}, \bibinfo{author}{Hou, Y.}, \bibinfo{author}{Qiao,
  Y.}, \bibinfo{author}{Liu, Z.}, \bibinfo{year}{2023}.
\newblock \bibinfo{title}{Rethinking range view representation for lidar
  segmentation}, in: \bibinfo{booktitle}{Proc.~of the IEEE/CVF Intl.~Conf.~on
  Computer Vision (ICCV)}, pp. \bibinfo{pages}{228--240}.
\bibitem[{Lai et~al.(2023)Lai, Chen, Lu, Liu and Jia}]{lai2023spherical}
\bibinfo{author}{Lai, X.}, \bibinfo{author}{Chen, Y.}, \bibinfo{author}{Lu,
  F.}, \bibinfo{author}{Liu, J.}, \bibinfo{author}{Jia, J.},
  \bibinfo{year}{2023}.
\newblock \bibinfo{title}{Spherical transformer for lidar-based 3d
  recognition}, in: \bibinfo{booktitle}{Proc.~of the IEEE/CVF Conf.~on Computer
  Vision and Pattern Recognition (CVPR)}, pp. \bibinfo{pages}{17545--17555}.
\bibitem[{Li et~al.(2023)Li, Casas and Urtasun}]{li2023memoryseg}
\bibinfo{author}{Li, E.}, \bibinfo{author}{Casas, S.},
  \bibinfo{author}{Urtasun, R.}, \bibinfo{year}{2023}.
\newblock \bibinfo{title}{Memoryseg: Online lidar semantic segmentation with a
  latent memory}, in: \bibinfo{booktitle}{Proc.~of the IEEE/CVF Intl.~Conf.~on
  Computer Vision (ICCV)}, pp. \bibinfo{pages}{745--754}.
\bibitem[{Li et~al.(2022)Li, Dai and Ding}]{li2022self}
\bibinfo{author}{Li, J.}, \bibinfo{author}{Dai, H.}, \bibinfo{author}{Ding,
  Y.}, \bibinfo{year}{2022}.
\newblock \bibinfo{title}{Self-distillation for robust lidar semantic
  segmentation in autonomous driving}, in: \bibinfo{booktitle}{Proc.~of the
  Europ.~Conf.~on Computer Vision (ECCV)}, pp. \bibinfo{pages}{659--676}.
\bibitem[{Li et~al.(2025)Li, Shum and Breckon}]{li2025rapid}
\bibinfo{author}{Li, L.}, \bibinfo{author}{Shum, H.P.},
  \bibinfo{author}{Breckon, T.P.}, \bibinfo{year}{2025}.
\newblock \bibinfo{title}{Rapid-seg: Range-aware pointwise distance
  distribution networks for 3d lidar segmentation}, in:
  \bibinfo{booktitle}{Proc.~of the Europ.~Conf.~on Computer Vision (ECCV)}, pp.
  \bibinfo{pages}{222--241}.
\bibitem[{Li et~al.(2021)Li, Chen, Liu, Dai, Stachniss and Gall}]{li2021multi}
\bibinfo{author}{Li, S.}, \bibinfo{author}{Chen, X.}, \bibinfo{author}{Liu,
  Y.}, \bibinfo{author}{Dai, D.}, \bibinfo{author}{Stachniss, C.},
  \bibinfo{author}{Gall, J.}, \bibinfo{year}{2021}.
\newblock \bibinfo{title}{Multi-scale interaction for real-time lidar data
  segmentation on an embedded platform}.
\newblock \bibinfo{journal}{IEEE Robotics and Automation Letters (RA-L)}
  \bibinfo{volume}{7}, \bibinfo{pages}{738--745}.
\bibitem[{Liu et~al.(2023)Liu, Chen, Li, Kong, Yang, Xia, Bai, Zhu, Ma, Li
  et~al.}]{liu2023uniseg}
\bibinfo{author}{Liu, Y.}, \bibinfo{author}{Chen, R.}, \bibinfo{author}{Li,
  X.}, \bibinfo{author}{Kong, L.}, \bibinfo{author}{Yang, Y.},
  \bibinfo{author}{Xia, Z.}, \bibinfo{author}{Bai, Y.}, \bibinfo{author}{Zhu,
  X.}, \bibinfo{author}{Ma, Y.}, \bibinfo{author}{Li, Y.}, et~al.,
  \bibinfo{year}{2023}.
\newblock \bibinfo{title}{Uniseg: A unified multi-modal lidar segmentation
  network and the openpcseg codebase}, in: \bibinfo{booktitle}{Proc.~of the
  IEEE/CVF Intl.~Conf.~on Computer Vision (ICCV)}, pp.
  \bibinfo{pages}{21662--21673}.
\bibitem[{Meyer et~al.(2019)Meyer, Charland, Hegde, Laddha and
  Vallespi-Gonzalez}]{meyer2019sensor}
\bibinfo{author}{Meyer, G.P.}, \bibinfo{author}{Charland, J.},
  \bibinfo{author}{Hegde, D.}, \bibinfo{author}{Laddha, A.},
  \bibinfo{author}{Vallespi-Gonzalez, C.}, \bibinfo{year}{2019}.
\newblock \bibinfo{title}{Sensor fusion for joint 3d object detection and
  semantic segmentation}, in: \bibinfo{booktitle}{Proc.~of the IEEE/CVF Conf.
  on Computer Vision and Pattern Recognition Workshops}, pp.
  \bibinfo{pages}{0--0}.
\bibitem[{Milioto et~al.(2019)Milioto, Vizzo, Behley and
  Stachniss}]{milioto2019rangenet++}
\bibinfo{author}{Milioto, A.}, \bibinfo{author}{Vizzo, I.},
  \bibinfo{author}{Behley, J.}, \bibinfo{author}{Stachniss, C.},
  \bibinfo{year}{2019}.
\newblock \bibinfo{title}{Rangenet++: Fast and accurate lidar semantic
  segmentation}, in: \bibinfo{booktitle}{Proc.~of the IEEE/RSJ Intl.~Conf.~on
  Intelligent Robots and Systems (IROS)}, pp. \bibinfo{pages}{4213--4220}.
\bibitem[{Nisar and Waslander(2025)}]{nisar2025psa}
\bibinfo{author}{Nisar, B.}, \bibinfo{author}{Waslander, S.L.},
  \bibinfo{year}{2025}.
\newblock \bibinfo{title}{Psa-ssl: Pose and size-aware self-supervised learning
  on lidar point clouds}.
\newblock \bibinfo{journal}{arXiv preprint} .
\bibitem[{Nunes et~al.(2022)Nunes, Marcuzzi, Chen, Behley and
  Stachniss}]{nunes2022segcontrast}
\bibinfo{author}{Nunes, L.}, \bibinfo{author}{Marcuzzi, R.},
  \bibinfo{author}{Chen, X.}, \bibinfo{author}{Behley, J.},
  \bibinfo{author}{Stachniss, C.}, \bibinfo{year}{2022}.
\newblock \bibinfo{title}{Segcontrast: 3d point cloud feature representation
  learning through self-supervised segment discrimination}.
\newblock \bibinfo{journal}{IEEE Robotics and Automation Letters (RA-L)}
  \bibinfo{volume}{7}, \bibinfo{pages}{2116--2123}.
\bibitem[{Puy et~al.(2023)Puy, Boulch and Marlet}]{puy2023using}
\bibinfo{author}{Puy, G.}, \bibinfo{author}{Boulch, A.},
  \bibinfo{author}{Marlet, R.}, \bibinfo{year}{2023}.
\newblock \bibinfo{title}{Using a waffle iron for automotive point cloud
  semantic segmentation}, in: \bibinfo{booktitle}{Proc.~of the IEEE/CVF
  Intl.~Conf.~on Computer Vision (ICCV)}, pp. \bibinfo{pages}{3379--3389}.
\bibitem[{Qi et~al.(2017a)Qi, Su, Mo and Guibas}]{qi2017pointnet}
\bibinfo{author}{Qi, C.}, \bibinfo{author}{Su, H.}, \bibinfo{author}{Mo, K.},
  \bibinfo{author}{Guibas, L.J.}, \bibinfo{year}{2017}a.
\newblock \bibinfo{title}{Pointnet: Deep learning on point sets for 3d
  classification and segmentation}.
\newblock \bibinfo{journal}{Proc.~of the IEEE/CVF Conf.~on Computer Vision and
  Pattern Recognition (CVPR)} , \bibinfo{pages}{77--85}.
\bibitem[{Qi et~al.(2017b)Qi, Yi, Su and Guibas}]{qi2017pointnet++}
\bibinfo{author}{Qi, C.R.}, \bibinfo{author}{Yi, L.}, \bibinfo{author}{Su, H.},
  \bibinfo{author}{Guibas, L.J.}, \bibinfo{year}{2017}b.
\newblock \bibinfo{title}{Pointnet++: Deep hierarchical feature learning on
  point sets in a metric space}.
\newblock \bibinfo{journal}{Proc.~of the Conf.~on Neural Information Processing
  Systems (NeurIPS)} \bibinfo{volume}{30}.
\bibitem[{Razani et~al.(2021)Razani, Cheng, Taghavi and
  Bingbing}]{razani2021lite}
\bibinfo{author}{Razani, R.}, \bibinfo{author}{Cheng, R.},
  \bibinfo{author}{Taghavi, E.}, \bibinfo{author}{Bingbing, L.},
  \bibinfo{year}{2021}.
\newblock \bibinfo{title}{Lite-hdseg: Lidar semantic segmentation using lite
  harmonic dense convolutions}, in: \bibinfo{booktitle}{Proc.~of the IEEE
  Intl.~Conf.~on Robotics \& Automation (ICRA)}, pp.
  \bibinfo{pages}{9550--9556}.
\bibitem[{Schieber et~al.(2022)Schieber, Duerr, Schoen and
  Beyerer}]{schieber2022deep}
\bibinfo{author}{Schieber, H.}, \bibinfo{author}{Duerr, F.},
  \bibinfo{author}{Schoen, T.}, \bibinfo{author}{Beyerer, J.},
  \bibinfo{year}{2022}.
\newblock \bibinfo{title}{Deep sensor fusion with pyramid fusion networks for
  3d semantic segmentation}, in: \bibinfo{booktitle}{Proc.~of the IEEE Vehicles
  Symposium (IV)}, pp. \bibinfo{pages}{375--381}.
\bibitem[{Sun et~al.(2023)Sun, Dong, Huang, Ma, Xia, Xue, Wang and
  Wei}]{sun2023retentive}
\bibinfo{author}{Sun, Y.}, \bibinfo{author}{Dong, L.}, \bibinfo{author}{Huang,
  S.}, \bibinfo{author}{Ma, S.}, \bibinfo{author}{Xia, Y.},
  \bibinfo{author}{Xue, J.}, \bibinfo{author}{Wang, J.}, \bibinfo{author}{Wei,
  F.}, \bibinfo{year}{2023}.
\newblock \bibinfo{title}{Retentive network: A successor to transformer for
  large language models}.
\newblock \bibinfo{journal}{arXiv preprint} .
\bibitem[{Tang et~al.(2020)Tang, Liu, Zhao, Lin, Lin, Wang and
  Han}]{tang2020searching}
\bibinfo{author}{Tang, H.}, \bibinfo{author}{Liu, Z.}, \bibinfo{author}{Zhao,
  S.}, \bibinfo{author}{Lin, Y.}, \bibinfo{author}{Lin, J.},
  \bibinfo{author}{Wang, H.}, \bibinfo{author}{Han, S.}, \bibinfo{year}{2020}.
\newblock \bibinfo{title}{Searching efficient 3d architectures with sparse
  point-voxel convolution}, in: \bibinfo{booktitle}{Proc.~of the
  Europ.~Conf.~on Computer Vision (ECCV)}, pp. \bibinfo{pages}{685--702}.
\bibitem[{Tchapmi et~al.(2017)Tchapmi, Choy, Armeni, Gwak and
  Savarese}]{tchapmi2017segcloud}
\bibinfo{author}{Tchapmi, L.}, \bibinfo{author}{Choy, C.},
  \bibinfo{author}{Armeni, I.}, \bibinfo{author}{Gwak, J.},
  \bibinfo{author}{Savarese, S.}, \bibinfo{year}{2017}.
\newblock \bibinfo{title}{Segcloud: Semantic segmentation of 3d point clouds},
  in: \bibinfo{booktitle}{Proc.~of the Intl.~Conf.~on 3D Vision (3DV)}, pp.
  \bibinfo{pages}{537--547}.
\bibitem[{Thomas et~al.(2019)Thomas, Qi, Deschaud, Marcotegui, Goulette and
  Guibas}]{thomas2019kpconv}
\bibinfo{author}{Thomas, H.}, \bibinfo{author}{Qi, C.R.},
  \bibinfo{author}{Deschaud, J.E.}, \bibinfo{author}{Marcotegui, B.},
  \bibinfo{author}{Goulette, F.}, \bibinfo{author}{Guibas, L.J.},
  \bibinfo{year}{2019}.
\newblock \bibinfo{title}{Kpconv: Flexible and deformable convolution for point
  clouds}, in: \bibinfo{booktitle}{Proc.~of the IEEE/CVF Intl.~Conf.~on
  Computer Vision (ICCV)}, pp. \bibinfo{pages}{6411--6420}.
\bibitem[{Vaswani et~al.(2017)Vaswani, Shazeer, Parmar, Uszkoreit, Jones,
  Gomez, Kaiser and Polosukhin}]{vaswani2017attention}
\bibinfo{author}{Vaswani, A.}, \bibinfo{author}{Shazeer, N.},
  \bibinfo{author}{Parmar, N.}, \bibinfo{author}{Uszkoreit, J.},
  \bibinfo{author}{Jones, L.}, \bibinfo{author}{Gomez, A.N.},
  \bibinfo{author}{Kaiser, {\L}.}, \bibinfo{author}{Polosukhin, I.},
  \bibinfo{year}{2017}.
\newblock \bibinfo{title}{Attention is all you need}.
\newblock \bibinfo{journal}{Proc.~of the Conf.~on Neural Information Processing
  Systems (NeurIPS)} \bibinfo{volume}{30}.
\bibitem[{Wang et~al.(2021)Wang, Xie, Li, Fan, Song, Liang, Lu, Luo and
  Shao}]{wang2021pyramid}
\bibinfo{author}{Wang, W.}, \bibinfo{author}{Xie, E.}, \bibinfo{author}{Li,
  X.}, \bibinfo{author}{Fan, D.P.}, \bibinfo{author}{Song, K.},
  \bibinfo{author}{Liang, D.}, \bibinfo{author}{Lu, T.}, \bibinfo{author}{Luo,
  P.}, \bibinfo{author}{Shao, L.}, \bibinfo{year}{2021}.
\newblock \bibinfo{title}{Pyramid vision transformer: A versatile backbone for
  dense prediction without convolutions}, in: \bibinfo{booktitle}{Proc.~of the
  IEEE/CVF Intl.~Conf.~on Computer Vision (ICCV)}, pp.
  \bibinfo{pages}{568--578}.
\bibitem[{Wang et~al.(2019)Wang, Sun, Liu, Sarma, Bronstein and
  Solomon}]{wang2019dynamic}
\bibinfo{author}{Wang, Y.}, \bibinfo{author}{Sun, Y.}, \bibinfo{author}{Liu,
  Z.}, \bibinfo{author}{Sarma, S.E.}, \bibinfo{author}{Bronstein, M.M.},
  \bibinfo{author}{Solomon, J.M.}, \bibinfo{year}{2019}.
\newblock \bibinfo{title}{Dynamic graph cnn for learning on point clouds}.
\newblock \bibinfo{journal}{ACM Trans.~on Graphics} \bibinfo{volume}{38},
  \bibinfo{pages}{1--12}.
\bibitem[{Wu et~al.(2018)Wu, Wan, Yue and Keutzer}]{wu2018squeezeseg}
\bibinfo{author}{Wu, B.}, \bibinfo{author}{Wan, A.}, \bibinfo{author}{Yue, X.},
  \bibinfo{author}{Keutzer, K.}, \bibinfo{year}{2018}.
\newblock \bibinfo{title}{Squeezeseg: Convolutional neural nets with recurrent
  crf for real-time road-object segmentation from 3d lidar point cloud}, in:
  \bibinfo{booktitle}{Proc.~of the IEEE Intl.~Conf.~on Robotics \& Automation
  (ICRA)}, pp. \bibinfo{pages}{1887--1893}.
\bibitem[{Wu et~al.(2019a)Wu, Zhou, Zhao, Yue and Keutzer}]{wu2019squeezesegv2}
\bibinfo{author}{Wu, B.}, \bibinfo{author}{Zhou, X.}, \bibinfo{author}{Zhao,
  S.}, \bibinfo{author}{Yue, X.}, \bibinfo{author}{Keutzer, K.},
  \bibinfo{year}{2019}a.
\newblock \bibinfo{title}{Squeezesegv2: Improved model structure and
  unsupervised domain adaptation for road-object segmentation from a lidar
  point cloud}, in: \bibinfo{booktitle}{Proc.~of the IEEE Intl.~Conf.~on
  Robotics \& Automation (ICRA)}, pp. \bibinfo{pages}{4376--4382}.
\bibitem[{Wu et~al.(2019b)Wu, Qi and Fuxin}]{wu2019pointconv}
\bibinfo{author}{Wu, W.}, \bibinfo{author}{Qi, Z.}, \bibinfo{author}{Fuxin,
  L.}, \bibinfo{year}{2019}b.
\newblock \bibinfo{title}{Pointconv: Deep convolutional networks on 3d point
  clouds}, in: \bibinfo{booktitle}{Proc.~of the IEEE/CVF Conf.~on Computer
  Vision and Pattern Recognition (CVPR)}, pp. \bibinfo{pages}{9621--9630}.
\bibitem[{Wu et~al.(2024)Wu, Jiang, Wang, Liu, Liu, Qiao, Ouyang, He and
  Zhao}]{wu2024point}
\bibinfo{author}{Wu, X.}, \bibinfo{author}{Jiang, L.}, \bibinfo{author}{Wang,
  P.S.}, \bibinfo{author}{Liu, Z.}, \bibinfo{author}{Liu, X.},
  \bibinfo{author}{Qiao, Y.}, \bibinfo{author}{Ouyang, W.},
  \bibinfo{author}{He, T.}, \bibinfo{author}{Zhao, H.}, \bibinfo{year}{2024}.
\newblock \bibinfo{title}{Point transformer v3: Simpler faster stronger}, in:
  \bibinfo{booktitle}{Proc.~of the IEEE/CVF Conf.~on Computer Vision and
  Pattern Recognition (CVPR)}, pp. \bibinfo{pages}{4840--4851}.
\bibitem[{Wu et~al.(2022)Wu, Lao, Jiang, Liu and Zhao}]{wu2022point}
\bibinfo{author}{Wu, X.}, \bibinfo{author}{Lao, Y.}, \bibinfo{author}{Jiang,
  L.}, \bibinfo{author}{Liu, X.}, \bibinfo{author}{Zhao, H.},
  \bibinfo{year}{2022}.
\newblock \bibinfo{title}{Point transformer v2: Grouped vector attention and
  partition-based pooling}.
\newblock \bibinfo{journal}{Proc.~of the Conf.~on Neural Information Processing
  Systems (NeurIPS)} \bibinfo{volume}{35}, \bibinfo{pages}{33330--33342}.
\bibitem[{Xiao et~al.(2022)Xiao, Huang, Guan, Cui, Lu and
  Shao}]{xiao2022polarmix}
\bibinfo{author}{Xiao, A.}, \bibinfo{author}{Huang, J.}, \bibinfo{author}{Guan,
  D.}, \bibinfo{author}{Cui, K.}, \bibinfo{author}{Lu, S.},
  \bibinfo{author}{Shao, L.}, \bibinfo{year}{2022}.
\newblock \bibinfo{title}{Polarmix: A general data augmentation technique for
  lidar point clouds}.
\newblock \bibinfo{journal}{Proc.~of the Conf.~on Neural Information Processing
  Systems (NeurIPS)} \bibinfo{volume}{35}, \bibinfo{pages}{11035--11048}.
\bibitem[{Xiao et~al.(2021)Xiao, Shao, Hao, Zhang, Chai, Jiao, Li, Wu, Sun,
  Jiang et~al.}]{xiao2021pandaset}
\bibinfo{author}{Xiao, P.}, \bibinfo{author}{Shao, Z.}, \bibinfo{author}{Hao,
  S.}, \bibinfo{author}{Zhang, Z.}, \bibinfo{author}{Chai, X.},
  \bibinfo{author}{Jiao, J.}, \bibinfo{author}{Li, Z.}, \bibinfo{author}{Wu,
  J.}, \bibinfo{author}{Sun, K.}, \bibinfo{author}{Jiang, K.}, et~al.,
  \bibinfo{year}{2021}.
\newblock \bibinfo{title}{Pandaset: Advanced sensor suite dataset for
  autonomous driving}, in: \bibinfo{booktitle}{Proc.~of the IEEE Intl.~Conf.~on
  Intelligent Transportation Systems (ITSC)}, pp. \bibinfo{pages}{3095--3101}.
\bibitem[{Xu et~al.(2020)Xu, Wu, Wang, Zhan, Vajda, Keutzer and
  Tomizuka}]{xu2020squeezesegv3}
\bibinfo{author}{Xu, C.}, \bibinfo{author}{Wu, B.}, \bibinfo{author}{Wang, Z.},
  \bibinfo{author}{Zhan, W.}, \bibinfo{author}{Vajda, P.},
  \bibinfo{author}{Keutzer, K.}, \bibinfo{author}{Tomizuka, M.},
  \bibinfo{year}{2020}.
\newblock \bibinfo{title}{Squeezesegv3: Spatially-adaptive convolution for
  efficient point-cloud segmentation}, in: \bibinfo{booktitle}{Proc.~of the
  Europ.~Conf.~on Computer Vision (ECCV)}, pp. \bibinfo{pages}{1--19}.
\bibitem[{Xu et~al.(2021)Xu, Zhang, Dou, Zhu, Sun and Pu}]{xu2021rpvnet}
\bibinfo{author}{Xu, J.}, \bibinfo{author}{Zhang, R.}, \bibinfo{author}{Dou,
  J.}, \bibinfo{author}{Zhu, Y.}, \bibinfo{author}{Sun, J.},
  \bibinfo{author}{Pu, S.}, \bibinfo{year}{2021}.
\newblock \bibinfo{title}{Rpvnet: A deep and efficient range-point-voxel fusion
  network for lidar point cloud segmentation}, in: \bibinfo{booktitle}{Proc.~of
  the IEEE/CVF Intl.~Conf.~on Computer Vision (ICCV)}, pp.
  \bibinfo{pages}{16024--16033}.
\bibitem[{Xu et~al.(2023)Xu, Kong, Shuai and Liu}]{xu2023frnet}
\bibinfo{author}{Xu, X.}, \bibinfo{author}{Kong, L.}, \bibinfo{author}{Shuai,
  H.}, \bibinfo{author}{Liu, Q.}, \bibinfo{year}{2023}.
\newblock \bibinfo{title}{Frnet: Frustum-range networks for scalable lidar
  segmentation}.
\newblock \bibinfo{journal}{arXiv preprint} .
\bibitem[{Yan et~al.(2022)Yan, Gao, Zheng, Zheng, Zhang, Cui and
  Li}]{yan20222dpass}
\bibinfo{author}{Yan, X.}, \bibinfo{author}{Gao, J.}, \bibinfo{author}{Zheng,
  C.}, \bibinfo{author}{Zheng, C.}, \bibinfo{author}{Zhang, R.},
  \bibinfo{author}{Cui, S.}, \bibinfo{author}{Li, Z.}, \bibinfo{year}{2022}.
\newblock \bibinfo{title}{2dpass: 2d priors assisted semantic segmentation on
  lidar point clouds}, in: \bibinfo{booktitle}{Proc.~of the Europ.~Conf.~on
  Computer Vision (ECCV)}, pp. \bibinfo{pages}{677--695}.
\bibitem[{Yan et~al.(2020)Yan, Zheng, Li, Wang and Cui}]{yan2020pointasnl}
\bibinfo{author}{Yan, X.}, \bibinfo{author}{Zheng, C.}, \bibinfo{author}{Li,
  Z.}, \bibinfo{author}{Wang, S.}, \bibinfo{author}{Cui, S.},
  \bibinfo{year}{2020}.
\newblock \bibinfo{title}{Pointasnl: Robust point clouds processing using
  nonlocal neural networks with adaptive sampling}, in:
  \bibinfo{booktitle}{Proc.~of the IEEE/CVF Conf.~on Computer Vision and
  Pattern Recognition (CVPR)}, pp. \bibinfo{pages}{5589--5598}.
\bibitem[{Zhang et~al.(2020)Zhang, Zhou, David, Yue, Xi, Gong and
  Foroosh}]{zhang2020polarnet}
\bibinfo{author}{Zhang, Y.}, \bibinfo{author}{Zhou, Z.},
  \bibinfo{author}{David, P.}, \bibinfo{author}{Yue, X.}, \bibinfo{author}{Xi,
  Z.}, \bibinfo{author}{Gong, B.}, \bibinfo{author}{Foroosh, H.},
  \bibinfo{year}{2020}.
\newblock \bibinfo{title}{Polarnet: An improved grid representation for online
  lidar point clouds semantic segmentation}, in: \bibinfo{booktitle}{Proc.~of
  the IEEE/CVF Conf.~on Computer Vision and Pattern Recognition (CVPR)}, pp.
  \bibinfo{pages}{9601--9610}.
\bibitem[{Zhao et~al.(2021a)Zhao, Jiang, Jia, Torr and Koltun}]{zhao2021point}
\bibinfo{author}{Zhao, H.}, \bibinfo{author}{Jiang, L.}, \bibinfo{author}{Jia,
  J.}, \bibinfo{author}{Torr, P.H.}, \bibinfo{author}{Koltun, V.},
  \bibinfo{year}{2021}a.
\newblock \bibinfo{title}{Point transformer}, in: \bibinfo{booktitle}{Proc.~of
  the IEEE/CVF Intl.~Conf.~on Computer Vision (ICCV)}, pp.
  \bibinfo{pages}{16259--16268}.
\bibitem[{Zhao et~al.(2021b)Zhao, Bai and Huang}]{zhao2021fidnet}
\bibinfo{author}{Zhao, Y.}, \bibinfo{author}{Bai, L.}, \bibinfo{author}{Huang,
  X.}, \bibinfo{year}{2021}b.
\newblock \bibinfo{title}{Fidnet: Lidar point cloud semantic segmentation with
  fully interpolation decoding}, in: \bibinfo{booktitle}{Proc.~of the IEEE/RSJ
  Intl.~Conf.~on Intelligent Robots and Systems (IROS)}, pp.
  \bibinfo{pages}{4453--4458}.
\bibitem[{Zhu et~al.(2021)Zhu, Zhou, Wang, Hong, Ma, Li, Li and
  Lin}]{zhu2021cylindrical}
\bibinfo{author}{Zhu, X.}, \bibinfo{author}{Zhou, H.}, \bibinfo{author}{Wang,
  T.}, \bibinfo{author}{Hong, F.}, \bibinfo{author}{Ma, Y.},
  \bibinfo{author}{Li, W.}, \bibinfo{author}{Li, H.}, \bibinfo{author}{Lin,
  D.}, \bibinfo{year}{2021}.
\newblock \bibinfo{title}{Cylindrical and asymmetrical 3d convolution networks
  for lidar segmentation}, in: \bibinfo{booktitle}{Proc.~of the IEEE/CVF
  Conf.~on Computer Vision and Pattern Recognition (CVPR)}, pp.
  \bibinfo{pages}{9939--9948}.

\end{thebibliography}






\end{document}